\documentclass{article}

\usepackage{PRIMEarxiv}
\usepackage{amsmath, amsthm, amssymb}
\usepackage[utf8]{inputenc} 
\usepackage[T1]{fontenc}    
\usepackage{hyperref}       
\usepackage{url}            
\usepackage{booktabs}       
\usepackage{amsfonts}       
\usepackage{nicefrac}       
\usepackage{microtype}      
\usepackage{lipsum}
\usepackage{fancyhdr}       
\usepackage{graphicx}       
\graphicspath{{media/}}     
\usepackage{graphicx}
\usepackage{longtable}
\usepackage{siunitx}
\usepackage{multirow}
\usepackage{subfigure}
\usepackage{algorithm}
\usepackage{algorithmic}
\usepackage{cancel}
\usepackage{diagbox}
\graphicspath{{media/}}     

\title{Analysis of ODE2VAE with Examples
}
\author{
  Batuhan Koyuncu \\
  Computer Engineering Dept.\\
  Boğaziçi University \\
  İstanbul\\
  \texttt{batukoyuncu@gmail.com}
}

\begin{document}
\maketitle

\begin{abstract}
Deep generative models aim to learn underlying distributions that generate the observed data. Given the fact that the generative distribution may be complex and intractable, deep latent variable models use probabilistic frameworks to learn more expressive joint probability distributions over the data and their low-dimensional hidden variables. Learning complex probability distributions over sequential data without any supervision is a difficult task for deep generative models. Ordinary Differential Equation Variational Auto-Encoder (ODE2VAE) is a deep latent variable model that aims to learn complex distributions over high-dimensional sequential data and their low-dimensional representations. ODE2VAE infers continuous latent dynamics of the high-dimensional input in a low-dimensional hierarchical latent space. The hierarchical organization of the continuous latent space embeds a physics-guided inductive bias in the model. In this paper, we analyze the latent representations inferred by the ODE2VAE model over three different physical motion datasets: bouncing balls, projectile motion, and simple pendulum. Through our experiments, we explore the effects of the physics-guided inductive bias of the ODE2VAE model over the learned dynamical latent representations. We show that the model is able to learn meaningful latent representations to an extent without any supervision.
\end{abstract}


\section{Introduction}
The aim of generative modeling can be summarized as extracting features from the data and utilizing these features to approximate the underlying generative distribution \cite{DeepGenMod}. Deep generative models are composed of neural networks that parameterize the complex data distributions. High dimensional data such as sequences of images have an increased complexity which requires generative models to learn high-dimensional probability distributions that generate the data. 

Latent variable models propose a simple graphical model that introduces latent variables. The latent variables are assumed to be creating the data, but they are not observed. Therefore, latent variable models aim to learn an expressive joint distribution over the data points and latent variables. Deep latent variable models such as autoencoders (VAEs) \cite{Kingma2014variationalbayes} are parameterized by neural networks and trained in an unsupervised setting. Deep latent variable models may be used for data generation \cite{gp_vae}, anomaly  \cite{vae_anomaly}, representation learning \cite{vaerepresentationlearning, disentangled_rep_mandt}, time series forecasting \cite{vae_future,vae_mehrasa,vaesequential}, and learning continuous latent representations \cite{ode2vae,ode_cts_time_series,Rubanova2019,NCD_kidger2020,chen_neural_event}. Since the latent variables capture the hidden structure of the data, they may provide low-dimensional and interpretable representations which give insights about the true generative factors of the data. In order to learn meaningful latent variables, it is important to embed a proper inductive bias into deep generative models \cite{bengio_inductive,battaglia_inductive}.

Neural ordinary differential equations (Neural ODEs) \cite{Chen2018NODE} enable deep learning models to generate continuous latent representations by solving latent ordinary differential equations (ODEs). Moreover, Neural ODEs impose an inductive bias that facilitates learning continuous dynamics of physical systems \cite{Greydanus2019,hamiltonian_generative}. First order latent ODEs cannot capture non-linear effects over the latent representations, therefore, coupled latent ODEs may be used to capture non-linear dynamics such as collisions in the latent space \cite{Gwak2020}. Additionally, Neural ODEs can be trained in a Bayesian setting which may increase the model's robustness and quantify its uncertainty \cite{bayesian_node}.

ODE2VAE is a dynamic generative model that operates in continuous-time  \cite{ode2vae}. It aims to learn complex latent trajectories of sequential data by combining VAEs, Neural ODEs, and Bayesian neural networks (BNNs) \cite{gal2016bayesian}. The ODE2VAE model infers the latent trajectories by using coupled latent ODEs which induce a physics-motivated inductive bias into the model. In the original work, it is shown that the ODE2VAE model exceeds the performance of the compared models in the task of extrapolating future time steps of sequential datasets: CMU walking data, rotating MNIST, and bouncing balls datasets. However, the latent representations learned by the model are not explored. Given its physics-motivated inductive bias, the ODE2VAE model may learn meaningful latent representations which capture well-defined physical dynamics and approximate physical generating factors of a well-defined physical system. In order the explore these properties, the model can be challenged with different motion datasets to check if the learned dynamical representations follow physically meaningful trajectories. Our contributions are as follows:
\begin{itemize}
    \item We investigate ODE2VAE's performance on modeling the three motion datasets: bouncing balls, projectile motion, and simple pendulum.
     \item We uncover the effects of the inductive bias of the ODE2VAE model by analyzing the model's dynamical latent representations and their uncertainties.
\end{itemize}
This paper is derived from Batuhan Koyuncu's MS thesis "Analysis and Regularization of Deep Generative Second Order Ordinary Differential Equations" completed in 2021 at Boğaziçi University under the supervision of Prof. Lale Akarun.
\section{ODE2VAE model}
\label{sec:ode2vae}
ODE2VAE proposes a second order latent ODE model to infer latent representations of high-dimensional sequential data such as videos \cite{ode2vae}. The ODE2VAE model utilizes a continuous-time probabilistic latent ODE: it uses VAEs for encoding the input frames and Neural ODEs for modeling the latent trajectories. The latent trajectories correspond to the continuous latent representations of the sequential inputs inferred by the model \cite{ode2vae}. 

The ODE2VAE model with a second order latent ODE can formulated by using coupled first order latent ODEs. Therefore, the model's latent space can be decomposed into latent position, velocity, and acceleration trajectories. The latent position trajectory is formed by the latent representations that are used for decoding the data points. The latent velocity trajectory is consists of the latent representations which drive the dynamics of the position trajectory. Lastly, the latent acceleration trajectory corresponds to the latent acceleration field that drives the latent velocity trajectory. The hierarchy among the trajectories imposes the physics-guided intuitive bias of the model, since they model an arbitrary equation of motion in the latent space by using latent ODEs. The latent dynamics and underlying generative process for the sequential data $\mathbf{x}_{0:N}$ with $N+1$ frames at time points $0:T$ can be summarized as \cite{ode2vae}:
\begin{gather}
    \mathbf{s}_{0} \sim p(\mathbf{s}_{0})
\\
\mathbf{v}_{0} \sim p(\mathbf{v}_{0})
\\
\mathbf{s}_{t} = \mathbf{s}_{0} + \int_{0}^{t} \mathbf{v}_{\tau} d\tau
\\
\mathbf{v}_{t} = \mathbf{v}_{0} + \int_{0}^{t} \mathbf{f^{*}}(\mathbf{s}_{\tau},\mathbf{v}_{\tau}) d\tau
\\
\mathbf{x}_{i}\sim p(\mathbf{x}_{i} \mid \mathbf{s}_{i})
\end{gather}
where $p(\mathbf{s}_{0})$ and $p(\mathbf{v}_{0})$ denote true posterior distributions, $\mathbf{s}_{0}$ and $\mathbf{v}_{0}$ correspond to the initial latent states in the latent position and velocity spaces, $\mathbf{s}_{t} \in \mathbb{R}^{a}$ denotes the latent position at time $t$ and is driven by the differential field of $\mathbf{v}_{t}$, $\mathbf{v}_{t} \in \mathbb{R}^{a}$ denotes the latent velocity at time $t$ and is driven by the acceleration field $\mathbf{f^{*}}(\mathbf{s}_{\tau},\mathbf{v}_{\tau})$, and $p(\mathbf{x}_{i} \mid \mathbf{s}_{i})$ denotes the likelihood which is used to decode the frame $\mathbf{x}_{i}$. The ODE2VAE model optimizes the parameters of the generative model by using amortized variational inference methods \cite{Jordan1999,amortization}. The first components of the model's architecture are position and velocity encoders which are parameterized by convolutional neural networks:
\begin{align}
(\boldsymbol{\mu}_{\mathbf{s}}, \log \boldsymbol{\sigma}_{\mathbf{s}}) &=\mathcal{E}_{\texttt{pos}}(\mathbf{x}_{0})
\\
(\boldsymbol{\mu}_{\mathbf{v}}, \log \boldsymbol{\sigma}_{\mathbf{v}}) &=\mathcal{E}_{\texttt{vel}}(\mathbf{x}_{0:m})
\\
q_{\texttt{pos}}(\mathbf{s}_{0} \mid \mathbf{x}_{0}) &=\mathcal{N}(\mathbf{s}_{0} ; \boldsymbol{\mu}_{\mathbf{s}}, \operatorname{diag}(\boldsymbol{\sigma}_{\mathbf{s}}))
\label{}
\\
q_{\texttt{vel}}(\mathbf{v}_{0} \mid \mathbf{x}_{0}) &=\mathcal{N}(\mathbf{v}_{0} ; \boldsymbol{\mu}_{\mathbf{s}}, \operatorname{diag}(\boldsymbol{\sigma}_{\mathbf{v}}))
\label{}
\end{align}
where the approximate posterior distribution is denoted by $q$, the position encoder $\mathcal{E}_{\texttt{pos}}$ has the input of the first frame, the velocity encoder  $\mathcal{E}_{\texttt{vel}}$  has first $m$ frames as its input, $m$ is called amortized inference length. Both approximate posteriors follow multivariate Gaussian distributions parameterized by the mean and variance vectors generated by the corresponding encoders. The concatenation of the position and velocity latents can be denoted with a single random variable $\mathbf{z}_{t}=[\mathbf{s}_{t},\mathbf{v}_{t}]$. The amortized approximate posterior distribution for the initial latent state can be written as:
\begin{align}
    q_{\mathrm{enc}}\left(\mathbf{z}_{0}\mid\mathbf{x}_{0: N}\right)&=q_{\mathrm{enc}}\left(\left(\begin{array}{c}\mathbf{s}_{0} \\\mathbf{v}_{0}\end{array}\right) \mid \mathbf{x}_{0: N}\right)   \nonumber\\ &=\mathcal{N}\left(\left(\begin{array}{c}\boldsymbol{\mu}_{\mathbf{s}}(\mathbf{x}_{0}) \\ \boldsymbol{\mu}_{\mathbf{v}}(\mathbf{x}_{0:m})\end{array}\right),\left(\begin{array}{cc}\operatorname{diag}\left(\boldsymbol{\sigma}_{\mathbf{s}}(\mathbf{x}_{0})\right) & \mathbf{0} \\ \mathbf{0} & \operatorname{diag}\left(\boldsymbol{\sigma}_{\mathbf{v}}(\mathbf{x}_{0:m})\right)\end{array}\right)\right)
\end{align}
The following latent states can be model by using two first order coupled ODEs \cite{ode2vae}:
\begin{gather}
    \underbrace{\left[\begin{array}{l}\mathbf{s}_{t} \\ \mathbf{v}_{t}\end{array}\right]}_{\mathbf{z}_{t}}=\left[\begin{array}{l}\mathbf{s}_{0} \\ \mathbf{v}_{0}\end{array}\right]+\int_{0}^{t}\underbrace{\left[\begin{array}{c}\mathbf{v}_{\tau} \\ \mathbf{f}_{\mathcal{W}}\left(\mathbf{s}_{\tau}, \mathbf{v}_{\tau}\right)\end{array}\right]}_{\tilde{\mathbf{f}}_{\mathcal{W}}\left(\mathbf{s}_{\tau},\mathbf{v}_{\tau}\right)} d \tau
    \label{eq:2nd_order_ode}
    \\
    \dot{\mathbf{s}}_{t} = \mathbf{v}_{t} \qquad
    \dot{\mathbf{v}}_{t} = \mathbf{f}_{\mathcal{W}}\left(\mathbf{s}_{t}, \mathbf{v}_{t}\right)
\end{gather}
where $\mathbf{f}_{\mathcal{W}}(\mathbf{s}_{t},\mathbf{v}_{t})$ denotes the latent acceleration field which is parameterized by a BNN for approximating the true acceleration field. The variational posterior distribution over the BNN's parameters $\mathcal{W}$ is denoted as:

\begin{equation}
    q(\mathcal{W})= \mathcal{N}(\mathcal{W} ; \boldsymbol{\mu}_{\mathbf{f}}, \operatorname{diag}(\boldsymbol{\sigma}_{\mathbf{f}}))
\end{equation}

The differential field that governs logarithm of the approximate posterior for the latent states can be derived  using continuous normalizing flows \cite{Chen2018NODE} and instantaneous change of variables theorem following Eq. \ref{eq:2nd_order_ode}:
\begin{align}
    \frac{\partial \log q\left(\mathbf{z}_{t} \mid \mathcal{W}\right)}{\partial t}&=-\operatorname{Tr}\left(\frac{d {\tilde{\mathbf{f}}}_{\mathcal{W}}\left(\mathbf{z}_{t}\right)}{d \mathbf{z}_{t}}\right) d t
    \\
    &=-\operatorname{Tr}\left(\begin{array}{cc}\frac{\partial \mathbf{v}_{t}}{\partial \mathbf{s}_{t}} & \frac{\partial \mathbf{v}_{t}}{\partial \mathbf{v}_{t}} \\ \frac{\partial \mathbf{f}_{\mathcal{W}}\left(\mathbf{s}_{t}, \mathbf{v}_{t}\right)}{\partial \mathbf{s}_{t}} & \frac{\partial \mathbf{f}_{\mathcal{W}}\left(\mathbf{s}_{t}, \mathbf{v}_{t}\right)}{\partial \mathbf{v}_{t}}\end{array}\right)
    \\
    &=-\operatorname{Tr}\left(\frac{\partial \mathbf{f}_{\mathcal{W}}\left(\mathbf{s}_{t}, \mathbf{v}_{t}\right)}{\partial \mathbf{v}_{t}}\right)
\end{align}
Therefore, the variational log density over the dynamic latent states can be written by using continuous normalizing flows \cite{Chen2018NODE}:
\begin{equation}
\log q\left(\mathbf{z}_{T} \mid \mathcal{W}\right)=\log q\left(\mathbf{z}_{0} \mid \mathcal{W}\right)-\int_{0}^{T} \operatorname{Tr}\left(\frac{\partial \mathbf{f}_{\mathcal{W}}\left(\mathbf{s}_{\tau}, \mathbf{v}_{\tau}\right)}{\partial \mathbf{v}_{\tau}}\right) d \tau
\end{equation}
where $\mathbf{z}_{t}=[\mathbf{s}_{t},\mathbf{v}_{t}]$. The total approximate posterior can be factorized into the three components over the parameters of the BNN, encoder, and latent ODE as the following:
\begin{equation}
q\left(\mathcal{W}, \mathbf{z}_{0: N} \mid \mathbf{x}_{0:N}\right)=q(\mathcal{W}) q_{\text {enc}}\left(\mathbf{z}_{0} \mid \mathbf{x}_{0:N}\right) q_{\text {ode}}\left(\mathbf{z}_{1: N} \mid \mathbf{z}_{0}, \mathcal{W}\right)
\label{eq:factorized_approx_posterior}
\end{equation}

\begin{figure}[!htb]
	\begin{center}
		\includegraphics[width=0.70\columnwidth]{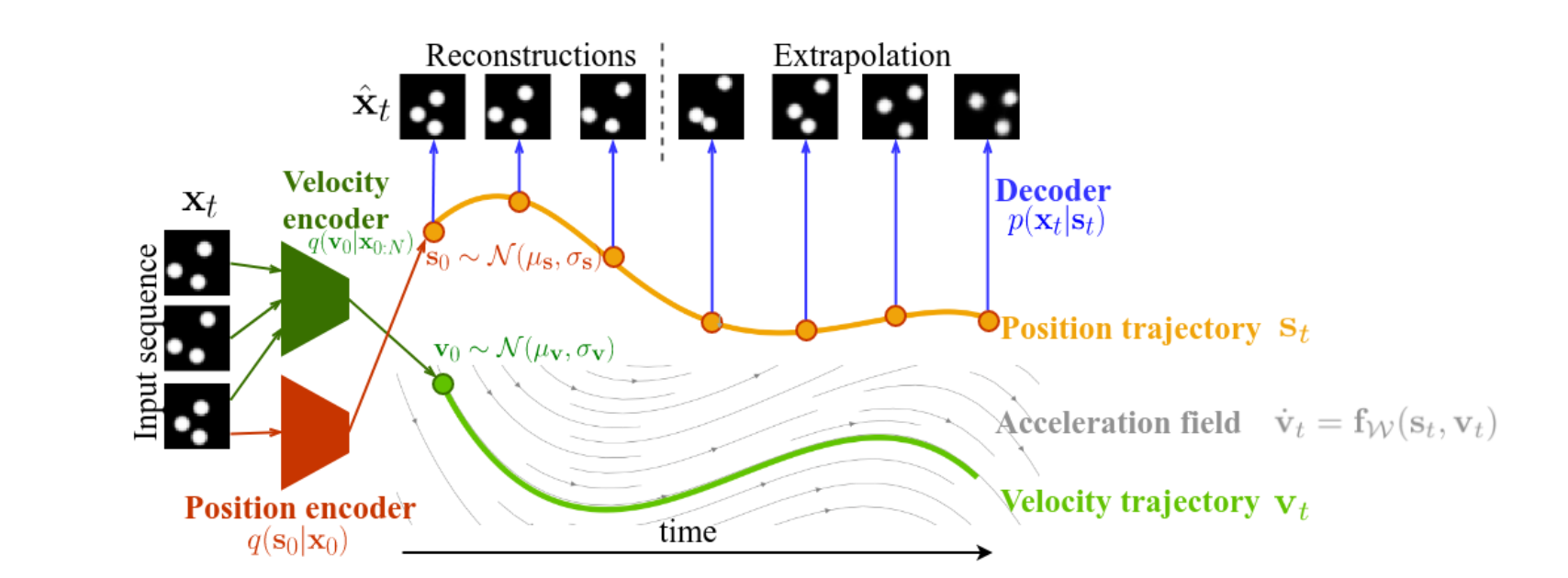}
		\caption[Illustration of the ODE2VAE model.]{Illustration of the ODE2VAE model. The figure is taken from \cite{ode2vae}.}
		\label{fig:ode2vae_scheme}
	\end{center}
\end{figure}

After the latent trajectories are inferred, the ODE2VAE model decodes the position latent variables using a decoder. The decoder is parameterized by a transposed convolutional neural network and it outputs parameters $\mathbf{p}$ for a factorized Bernoulli observation model \cite{ode2vae,Kingma2019introtoVAE}:
\begin{gather}
\mathbf{p} =\mathcal{D}_{\mathrm{pos}}(\mathbf{s_{t}}) 
\\
p(\mathbf{x}_{t} \mid \mathbf{s}_{t}) = \mathcal{B}(\mathbf{x}_{t};\mathbf{p})
\label{eq:ode2vae_decoder}
\end{gather}
where $\mathcal{D}$ denotes the decoder network and $\mathcal{B}(\mathbf{x};\mathbf{p})$ is a Bernoulli distribution over each pixel value given the fact that the data consist of grayscale images. The ELBO term, which is the lower bound of the marginal log-likelihood, can be written as  \cite{ode2vae}:
\begin{align}
    \log p(X) &\geq \underbrace{\mathcal{L}_{\mathrm{ODE2VAE}}(X)}_{\mathrm{ELBO}}
    \\&=\underbrace{-\mathcal{D}_{K L}[q(\mathcal{W}, Z \mid X)|| p(\mathcal{W}, Z)]}_{\mathrm{Regularization}}+\underbrace{\mathbb{E}_{q(\mathcal{W}, Z \mid X)}[\log p(X \mid \mathcal{W}, Z)]}_{\mathrm{Reconstruction}}
    \\
    &=-\underbrace{\mathcal{D}_{K L}[q(\mathcal{W}) \| p(\mathcal{W})]}_{\text {ODE regularization }}+\underbrace{\mathbb{E}_{q_{\mathrm{enc}}\left(\mathbf{z}_{0} \mid X\right)}\left[-\log \frac{q_{\mathrm{enc}}\left(\mathbf{z}_{0} \mid X\right)}{p\left(\mathbf{z}_{0}\right)}+\log p\left(\mathbf{x}_{0} \mid \mathbf{z}_{0}\right)\right]}_{\text {VAE loss }} \nonumber \\
    &\quad +\underbrace{\sum_{i=1}^{N}\mathbb{E}_{q_{\mathrm{enc}}\left(\mathbf{z}_{0} \mid X\right)}\left[ \mathbb{E}_{q_{\mathrm{ode}}\left(\mathcal{W}, \mathbf{z}_{i} \mid X, \mathbf{z}_{0}\right)}\left[-\log \frac{q_{\mathrm{ode}}\left(\mathbf{z}_{i} \mid \mathbf{z}_{0}, \mathcal{W}\right)}{p\left(\mathbf{z}_{i}\right)}+\log p\left(\mathbf{x}_{i} \mid \mathbf{z}_{i}\right)\right]\right]}_{\text {dynamic loss }}
    \label{eq:ode2vae_elbo_decomposed}
\end{align}
where $X$ and $Z$ denote the sequences $\mathbf{x}_{0:N}$ and $\mathbf{z}_{0:N}$. The priors $p(\mathcal{W})$, $p(\mathbf{z}_{0})$, and $p(\mathbf{z}_{1:N})$ are standard multivariate Gaussians. In Eq. \ref{eq:ode2vae_elbo_decomposed}, the first term regularizes the BNN by regularizing the network weights that model the latent acceleration field. The second term is in the form of a standard ELBO term of a VAE. The last term corresponds to an ELBO term for the latent sequence. The combined ELBO term is optimized with respect to weights of the BNN, parameters of the encoder and decoder networks by using mini-batches. Additionally, the baseline paper provides a penalized ELBO formalization that offers a more stable training of the model \cite{ode2vae}. The penalized version of the ELBO is:
\begin{align}
 \mathcal{L}_{\mathrm{ODE2VAE^{*}}}=&- \beta_{\mathcal{W}} \mathcal{D}_{K L}[q(\mathcal{W})|| p(\mathcal{W})]+\mathbb{E}_{q(\mathcal{W}, Z \mid X)}\left[-\log \frac{q(Z \mid \mathcal{W}, X)}{p(Z)}+\log p(X \mid \mathcal{W}, Z)\right] \nonumber
 \\ 
&-\gamma \mathbb{E}_{q(\mathcal{W})}\left[\mathcal{D}_{K L}\left[q_{\mathrm{ode}}(Z \mid \mathcal{W}, X) \| q_{\mathrm{enc}}(Z \mid X)\right]\right]
\end{align}
where $\beta_{W}=|a| /|\mathcal{W}|$ is chosen as the ratio of the latent space dimension $a$ and the number of weights in the BNN. It is noted that the penalized ELBO objective helps to balance the contributions of penalties over $\mathcal{W}$ and $\mathbf{z}_{i}$ \cite{ode2vae}. Additionally, it introduces another regularization term $\gamma$ which tunes the KL distance between the densities generated by the latent ODE and encoder to prevent the encoder from underfitting \cite{ode2vae}. The $\gamma$ term is chosen by cross-validation \cite{ode2vae}. In Figure \ref{fig:ode2vae_scheme}, one can find a illustrative scheme of the ODE2VAE model.

\section{Experiments}
In the experiments, three physical motion datasets are used. These are bouncing balls, simple pendulum, and projectile motion datasets. Compared to the baseline model \cite{ode2vae}, the datasets include two new variants of the bouncing balls dataset, and two new dynamical settings: simple pendulum and projectile motion. For each dynamical system, there are sequences of $32\times32\times1$ images with pixel values between 0 and 1.  For each dataset, the number of cases are $10000$, $500$, $500$ for training, validation, and test sets. Moreover, the moments of collisions and change of direction in the dynamics are stored and used in the analysis. We start the experiments with the minimum possible number of latent units for the given dataset and increase the dimensionality until the motion is captured. This approach creates a bottleneck in the latent dimensionality. It forces the model to learn meaningful latent representations because the ELBO objective can only be maximized by the latent units that represent ground truth generative factors.
All of the model variants are created by using the official ODE2VAE implementation \cite{ode2vae} and Tensorflow framework \cite{tensorflow2015-whitepaper}. The models are trained with Adam Optimizer \cite{adamKingmaB14} with the learning rate of $\eta=0.001$ and the batch size of 32. The amortized inference length is selected as $m=3$. All hyperparameters are chosen according to the baseline work \cite{ode2vae}. All experiments are executed on a single Tesla V100 GPU where each experiment takes approximately three days.
\subsection{Evaluation Metrics}
The experiment results are evaluated by using the following quantitative evaluation metrics. The quantitative evaluations are conducted by using the model's latent representations and its reconstructions and extrapolations. Through the experiments, we use the sample size $\mathrm{L}$=10, which is the number of latent states sampled at each time step.

\paragraph{L2 Norms of the Latent States} L2 norms of the latent states are used as explanatory and interpretable metrics used in the evaluation of the unsupervised deep generative models that operate with continuous-time data \cite{Rubanova2019,Gwak2020}. Since the baseline model operates with latent second order ODEs, it has the inductive bias of learning the latent dynamics similar to real motion dynamics. The L2 norm of the acceleration field is similar to the magnitude of the force effecting the dynamics in the latent space. Similarly, the L2 norm of the velocity latent variable resembles square root of the total kinetic energy of the system.

\paragraph{Mean Squared Error}
Mean squared error (MSE) computes mean of squared errors between true data points and predictions. It can be normalized with respect to the number of pixels in the data point, which is called pixel MSE. 
\paragraph{Peak Signal-to-Noise Ratio}
Peak signal-to-noise ratio (PSNR) measures the quality of the prediction by computing the logarithm of the ratio between the square of the maximum pixel fluctuation among the images and the pixel MSE between the ground truth and predicted image. As the PSNR score increases, the prediction quality also increases.
\paragraph{Negative Marginal Log-likelihood}
Negative marginal log-likelihood (NLL) of a data point under the variational model can be computed by using the importance sampling technique \cite{Rezende2014,Kingma2017thesis}.

\subsection{Bouncing Balls}
\label{sec:bouncing_balls}
\begin{figure}[!h]
\centering
\subfigure[Bouncing ball dataset with $n=1$.]{%
\includegraphics[width=.31\textwidth]{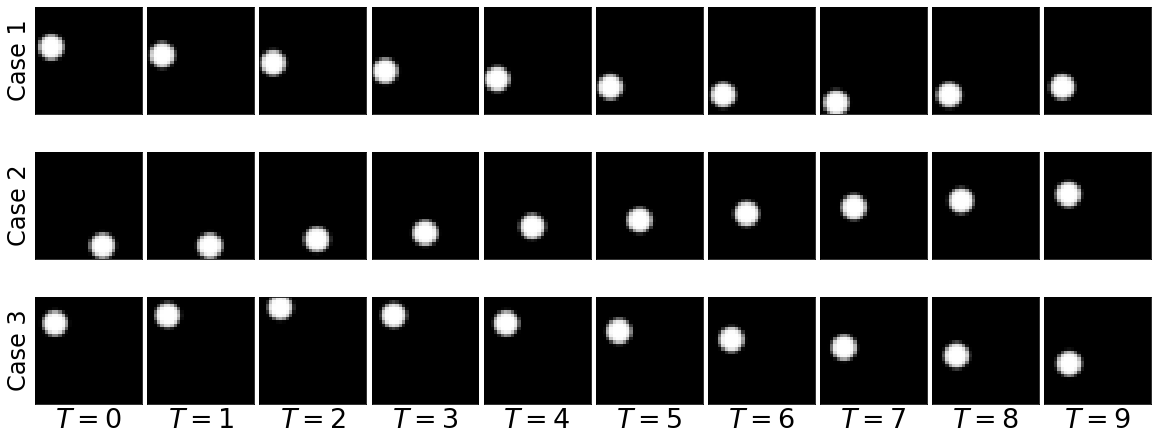}}%
\quad
\subfigure[Bouncing ball dataset with $n=2$.]{%
\includegraphics[width=.31\textwidth]{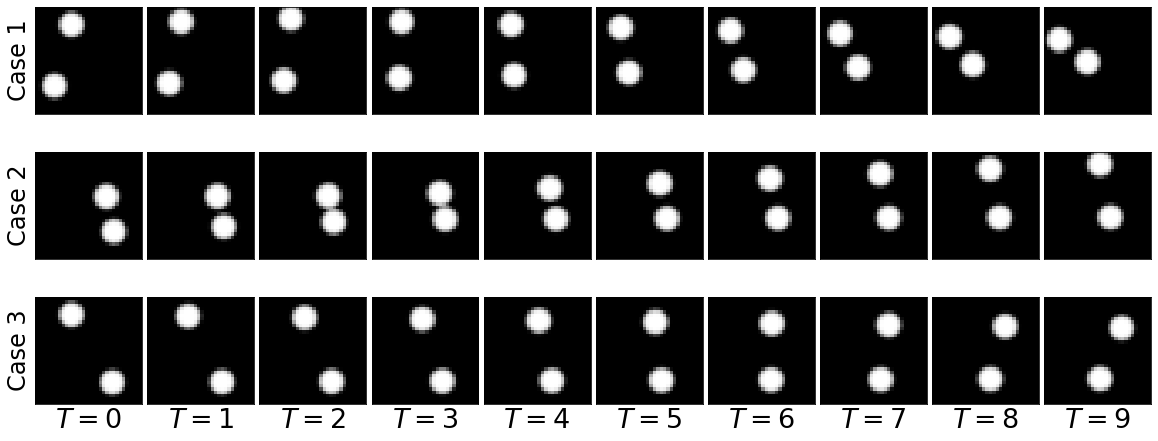}}%
\quad
\subfigure[Bouncing ball dataset with $n=3$.]{%
\includegraphics[width=.31\textwidth]{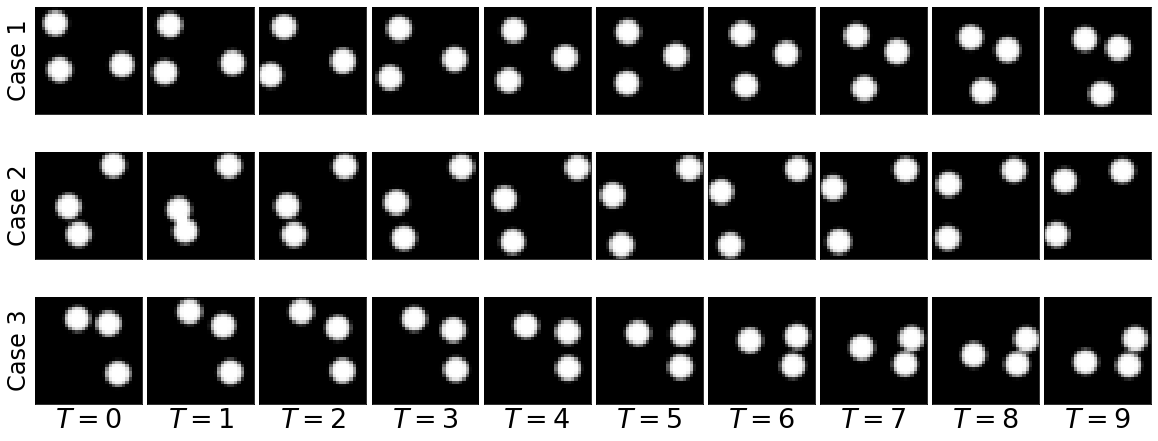}}%
\caption{Bouncing ball dataset with number of balls $n=1$, $n=2$, and $n=3$ with sequence length $T=10$.}
\label{fig:bball_n_123}
\end{figure}

Bouncing balls dataset is a standard benchmark task for models that aim to learn generative temporal modeling \cite{stanford_video_disentangle,ode2vae}. By using the provided implementation \cite{bball_first_paper}, we re-implemented the bouncing balls dataset with multiple variants. The datasets capture dynamics of $n$ balls in a 2D box. There is no friction in the motion and all collisions are elastic. The 2D box has a side length of $\SI{10.0}{\meter}$. The radius and mass of the balls are fixed, at $\SI{1.2}{\meter}$ and $\SI{1.0}{\kg}$, respectively. Each ball has a velocity that is randomly sampled from a standard normal distribution. The velocities of the balls in the same sequence are normalized, so that the total kinetic energy is constant over each sequence. The frames in the sequences are separated by one second, and the balls' motion is simulated with $0.5$ second resolution. In Figure \ref{fig:bball_n_123}, we present different variants of the bouncing ball datasets. We specify the sequence length as $T$ and the number of balls as $n$.

    \begin{table}[!h]
\caption[Performance metrics of the selected model on the single bouncing balls dataset.]{Performance metrics of the selected model on the bouncing balls dataset with $n=1$. For the MSE and NLL values, the lower is better. For the PSNR scores, the higher is better. Each metric is computed by using 10 samples per test case.}
\centering
\begin{tabular}{c *{3}{S[table-format=2,
                         table-column-width=7em]}
                }
    \toprule
            & \multicolumn{3}{c}{Metrics} 
                         \\
    \cmidrule(lr){2-4}
Model  & {MSE}    & {PSNR}    & {NLL}    \\
    \midrule
ODE2VAE, $a=3$       & {$0.0027\pm0.0032$}   & {$28.2601\pm4.6264$}    & {$42.5142$} \\
    \bottomrule
\end{tabular}
\label{table:bouncing_balls_n_1}
    \end{table}

\begin{figure}[!h]
\centering
\subfigure[MSE values for the bouncing balls dataset with $n=1$.]{%
\label{fig:n_1_mse}%
\includegraphics[height=1.73in]{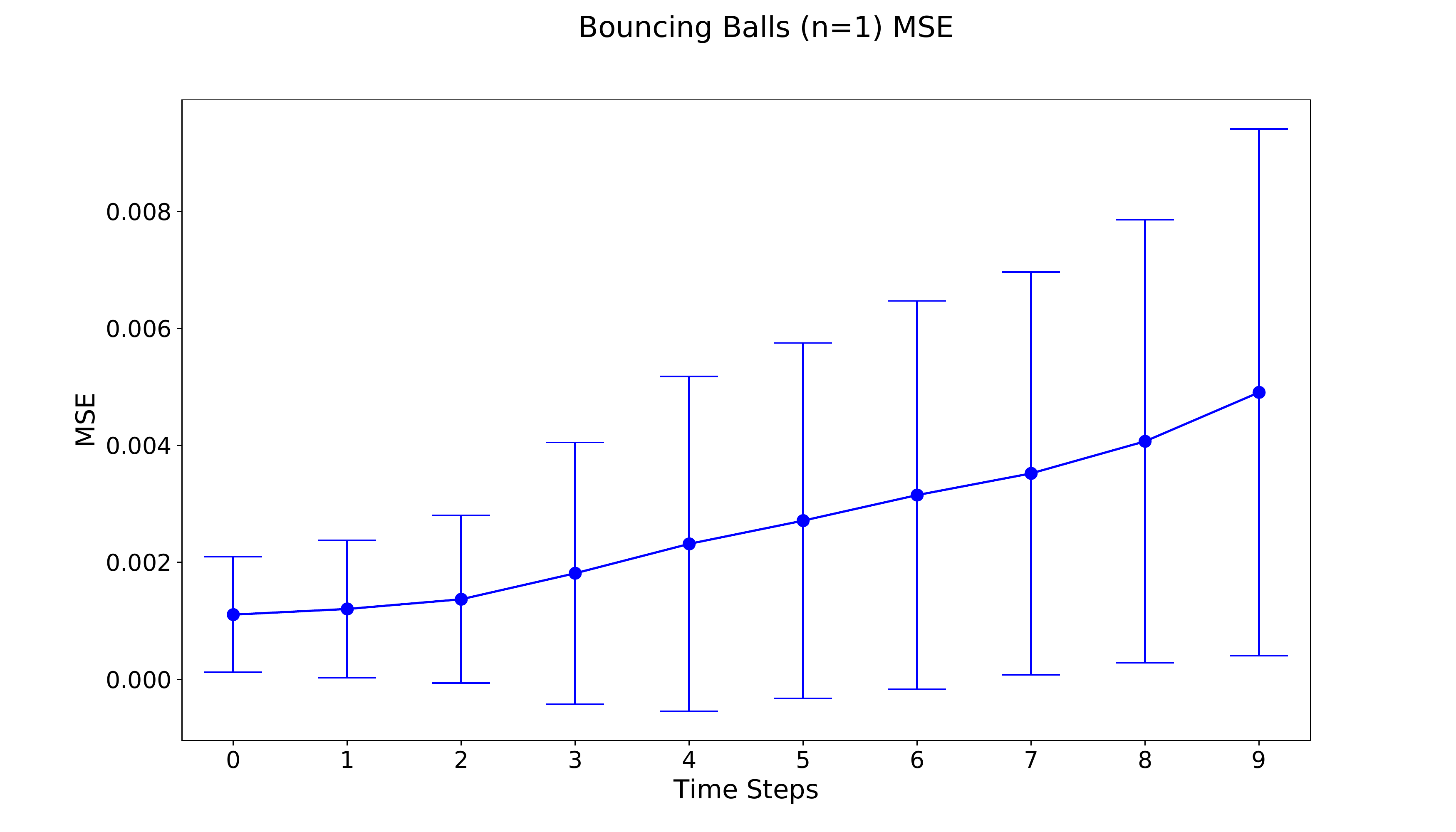}}%
\qquad
\subfigure[PSNR values for the bouncing balls dataset with $n=1$.]{%
\label{fig:n_1_psnr}%
\includegraphics[height=1.73in]{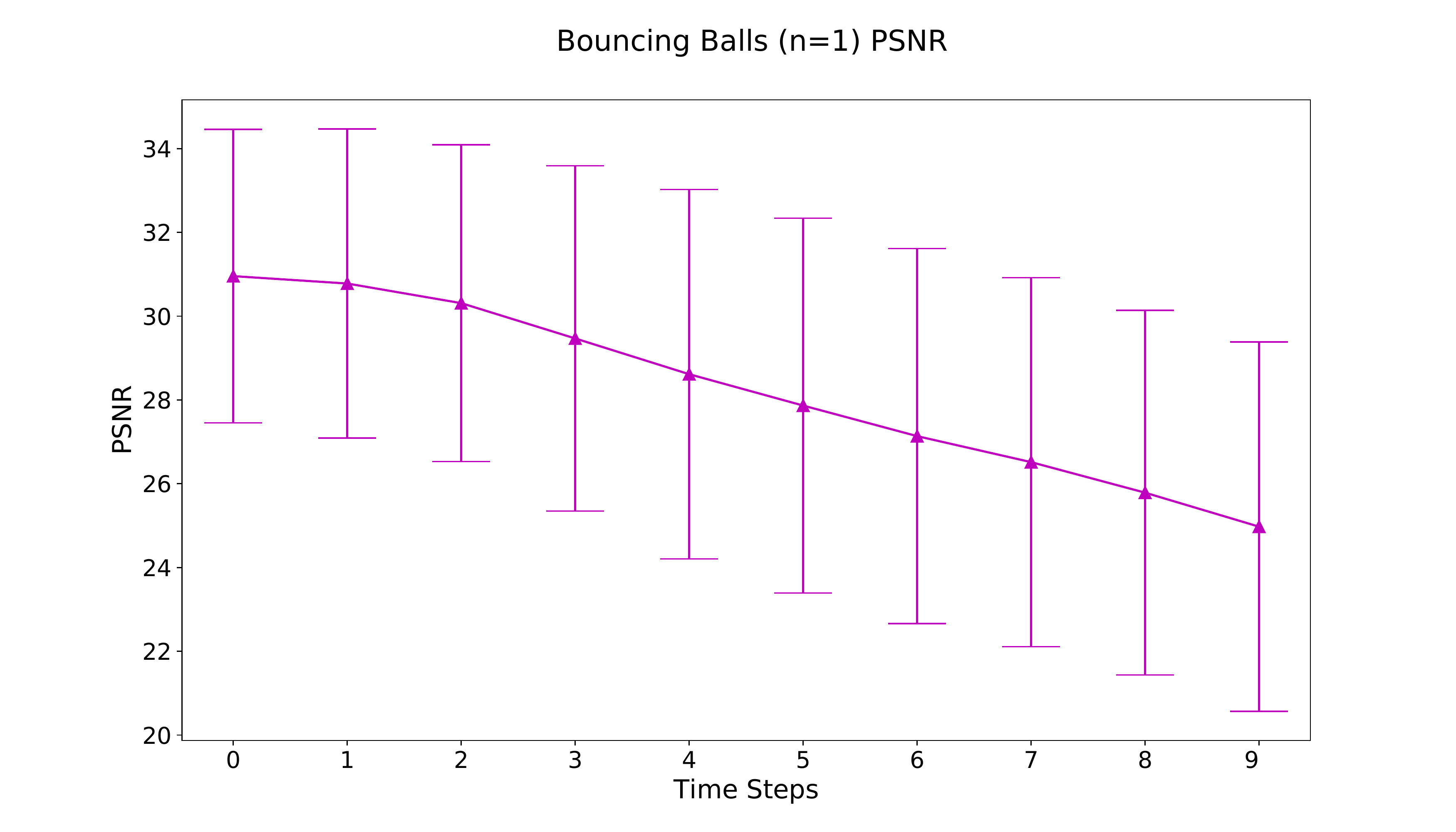}}%
\caption{MSE and PSNR values for the bouncing balls dataset with $n=1$.}
\label{fig:n_1_msepsnr}
\end{figure}

\begin{figure}[!h]
	\begin{center}
		\includegraphics[width=0.69\columnwidth]{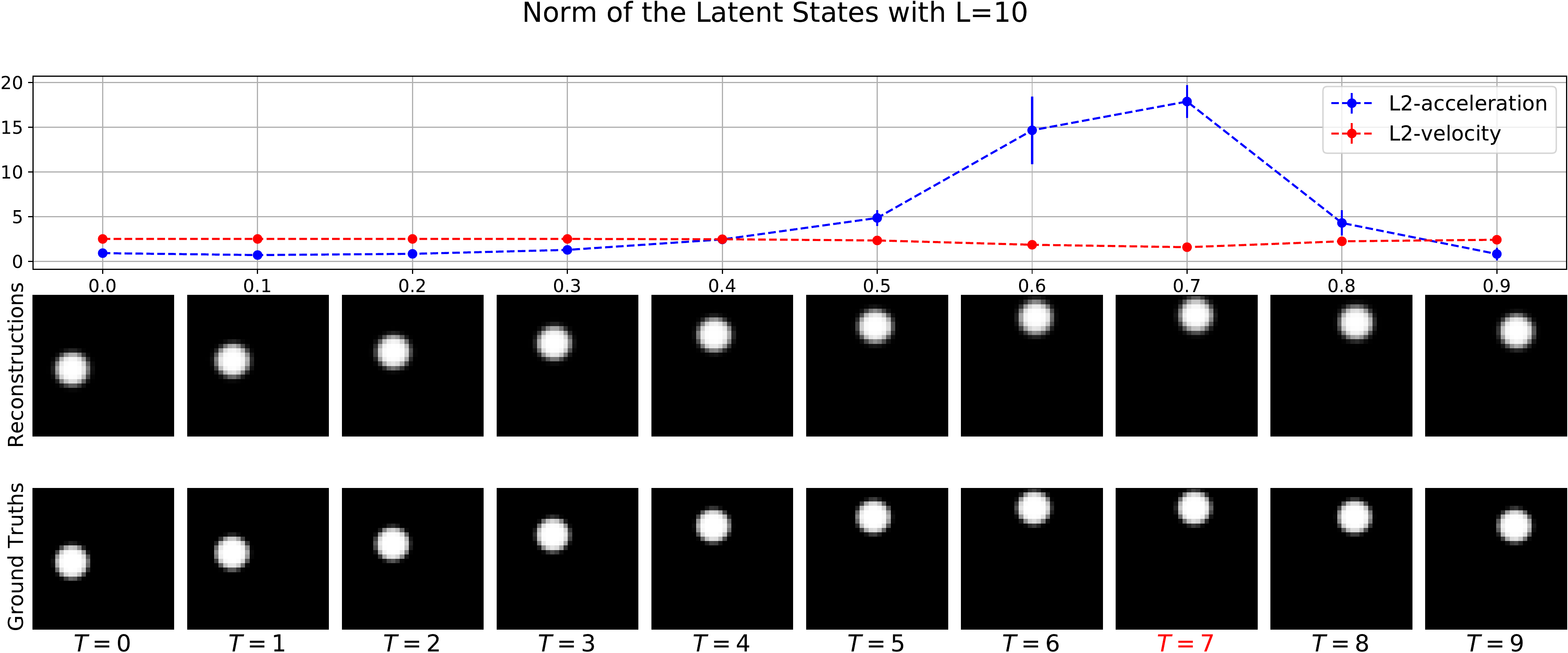}
		\caption[Example test case reconstructed by the ODE2VAE model for the bouncing ball motion with $n=1$.]{Example test case reconstructed by the ODE2VAE model with $a=3$, which is trained on the bouncing balls dataset with $n=1$. From  top  to  bottom: mean and standard deviation values of the latent norms; mean field prediction by the model with the sample size $\mathrm{L}$=10; ground truth frames. The indices of the collision times are highlighted.}
		\label{fig:n_1_figure}
	\end{center}
\end{figure}

\begin{figure}[!htb]
\centering
\subfigure[L2-norm of the Acceleration Latent]{%
\label{fig:n_1_norm_spreads}%
\includegraphics[height=1.68in]{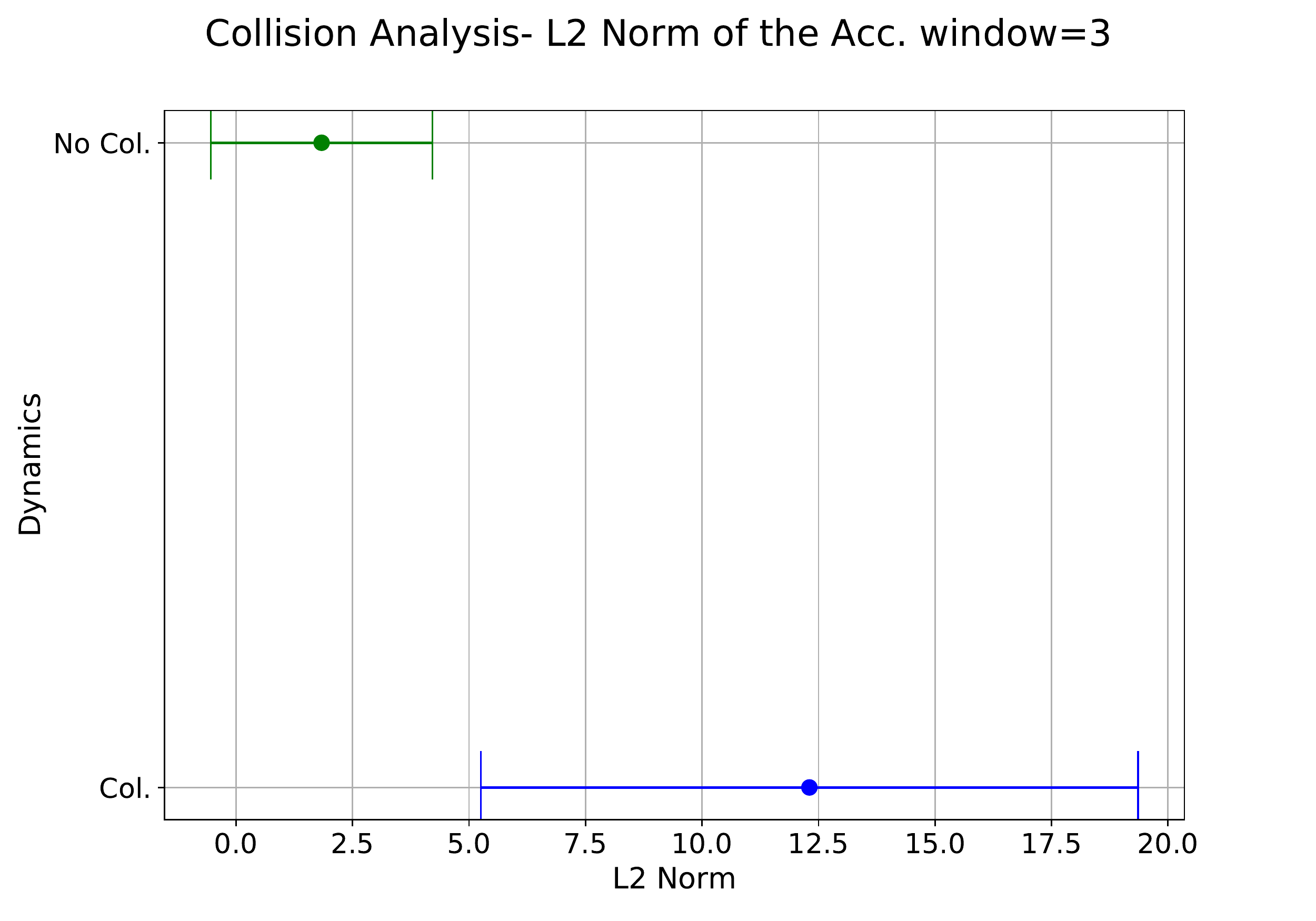}}%
\qquad
\subfigure[L2-norm of the Velocity Latent]{%
\label{fig:n_1_norm_spreads2}%
\includegraphics[height=1.68in]{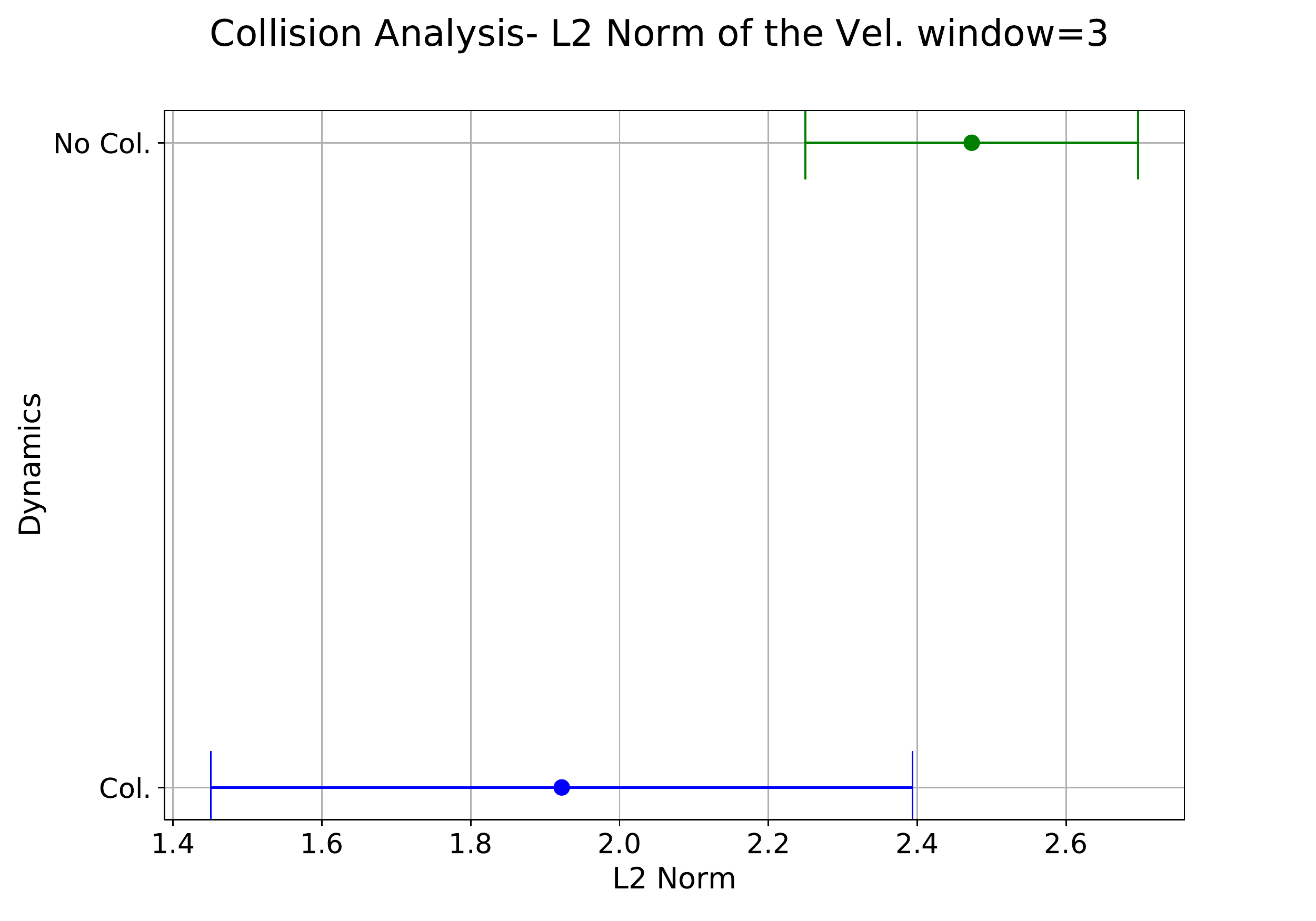}}%
\caption[Figure for the latent representation analysis of the baseline model for the bouncing balls dataset with $n=1$.]{Mean and standard deviation values for the L2-norm of the latent acceleration and velocity of the baseline model for the bouncing balls dataset with $n=1$. The figures display the values at the collision moments and the other time steps. The collision times are expanded with the window size of 3.}
\label{fig:n_1_norm_spread_all}
\end{figure}

\begin{figure}[!htb]
	\begin{center}
		\includegraphics[width=1\columnwidth]{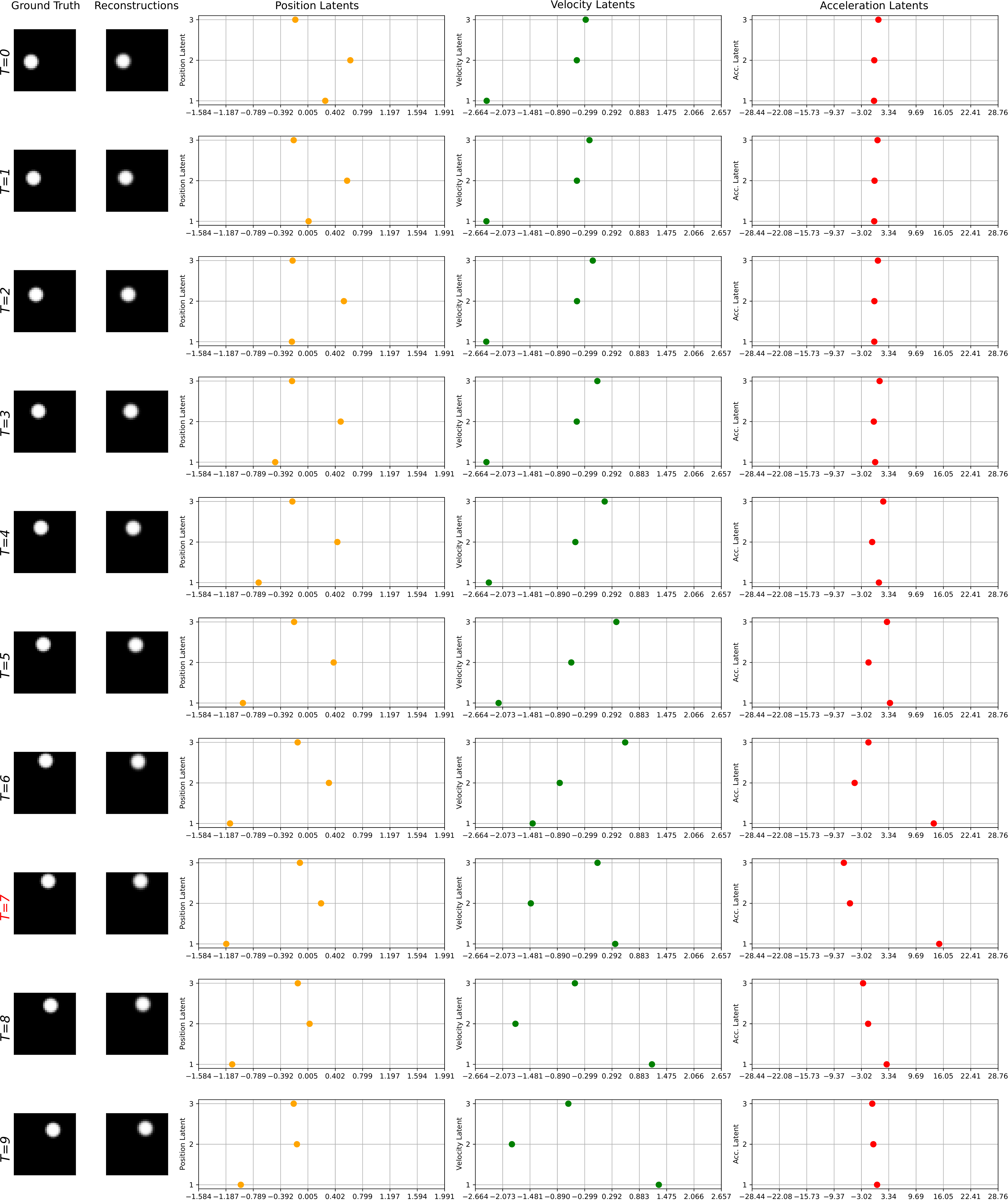}
		\caption[Example test case for bouncing ball with $n=1$ and corresponding latent vectors.]{Example test case for single bouncing ball with corresponding latent vectors. The figure explicitly shows the evolution of the latent vectors.}
		\label{fig:appendix_n_1}
	\end{center}
\end{figure}

The baseline model with the latent dimensionality $a=2$ cannot capture the dynamics of the motion of the single bouncing ball after it is trained for 250 epochs. We increase the latent dimensionality to $a=3$ and $a=12$ and train the models for 250 epochs. Both of the models have captured the dynamics of the single bouncing ball. In Table \ref{table:bouncing_balls_n_1}, we only report the metrics for model with $a=3$. In Figures \ref{fig:n_1_mse} and \ref{fig:n_1_psnr}, we plot MSE and PSNR values for the test cases over the time steps. The model is able to capture physically meaningful latent representations. We present an example case in Figure \ref{fig:n_1_figure}. When the ball hits the wall, there is a spike in the norm of the latent acceleration. It can be seen that the standard deviation of the norm of the acceleration field also increases during the collision. This indicates the fact that the output of the BNN has a greater uncertainty during the collision. Some possible reasons for the high uncertainty over the norm of the acceleration latents may be the non-linear motion during the collision and the scarcity of the time steps with collision. The norm of the latent velocity is not changed during the motion in the example case, which suggests the fact that the latent velocity preserves its norm, but changes the direction during the collisions (see Figure \ref{fig:appendix_n_1} for a detailed example). Figures \ref{fig:n_1_norm_spreads} and \ref{fig:n_1_norm_spreads2} summarize the statistics for the norm of the acceleration field and latent velocity over all test cases with a breakdown for the time steps with and without collision. Figure \ref{fig:n_1_norm_spreads} shows that the model generates an acceleration field with a greater magnitude during the collisions. At the collision moments, the high standard deviation over the norm of the acceleration field is meaningful since it depends on the amount of change in the momentum, which varies in the test cases. When there no collision, the model generates an acceleration field with a smaller norm compared to the mean magnitude at the collision time points, which is physically plausible. We note that the real dynamics require the model to generate zero acceleration field when there is no collision. Figure \ref{fig:n_1_norm_spreads2} shows that the model has increased the magnitude of the velocity latent without collision. Its magnitude decreases during the collision moments. This is not an expected behavior since the ball's stationary moments during the collisions are ignored in the dataset. Therefore, it can be said the model cannot preserve a constant norm of the latent velocity.

\newpage
    \begin{table}[!h]
\caption[Performance metrics of the selected model on the bouncing balls dataset with $n=2$.]{Performance metrics of the selected model on the bouncing balls dataset with $n=2$. For the MSE and NLL values, the lower is better. For the PSNR scores, the higher is better. Each metric is computed by using 10 samples per test case.}
\centering
\begin{tabular}{c *{3}{S[table-format=2,
                         table-column-width=7em]}
                }
    \toprule
            & \multicolumn{3}{c}{Metrics} 
                         \\
    \cmidrule(lr){2-4}
Model  & {MSE}    & {PSNR}    & {NLL}    \\
    \midrule
ODE2VAE, $a=5$       & {$0.0100\pm0.0123$}   & {$22.2161\pm4.3479$}    & {$96.4297$} \\
    \bottomrule
\end{tabular}
\label{table:bouncing_balls_n_2}
    \end{table}

\begin{figure}[!htb]
\centering
\subfigure[MSE values for the bouncing balls dataset with $n=2$.]{%
\includegraphics[height=1.73in]{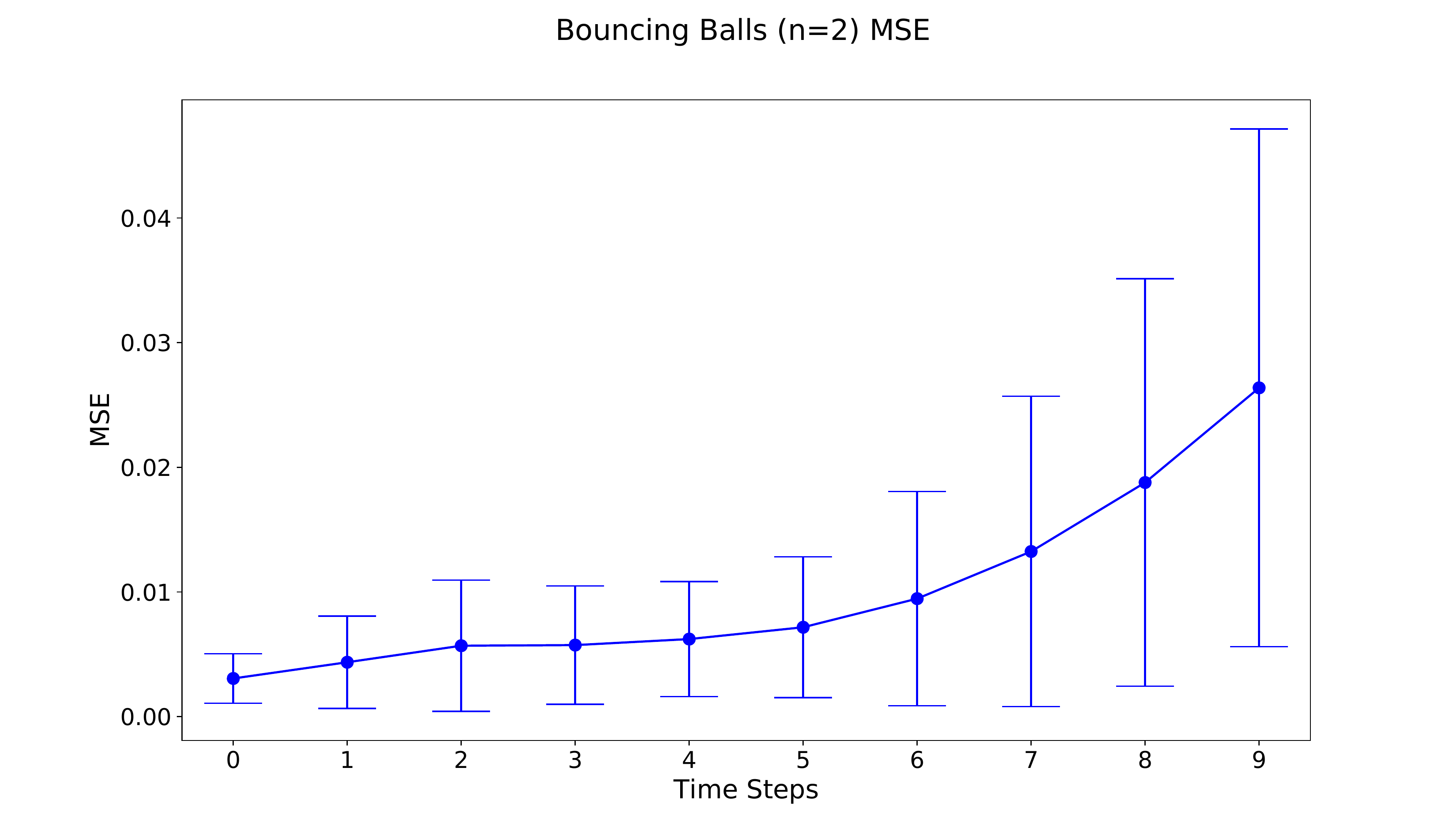}}
\label{fig:n_2_mse}
\qquad
\subfigure[PSNR values for the bouncing balls dataset with $n=2$.]{%
\includegraphics[height=1.73in]{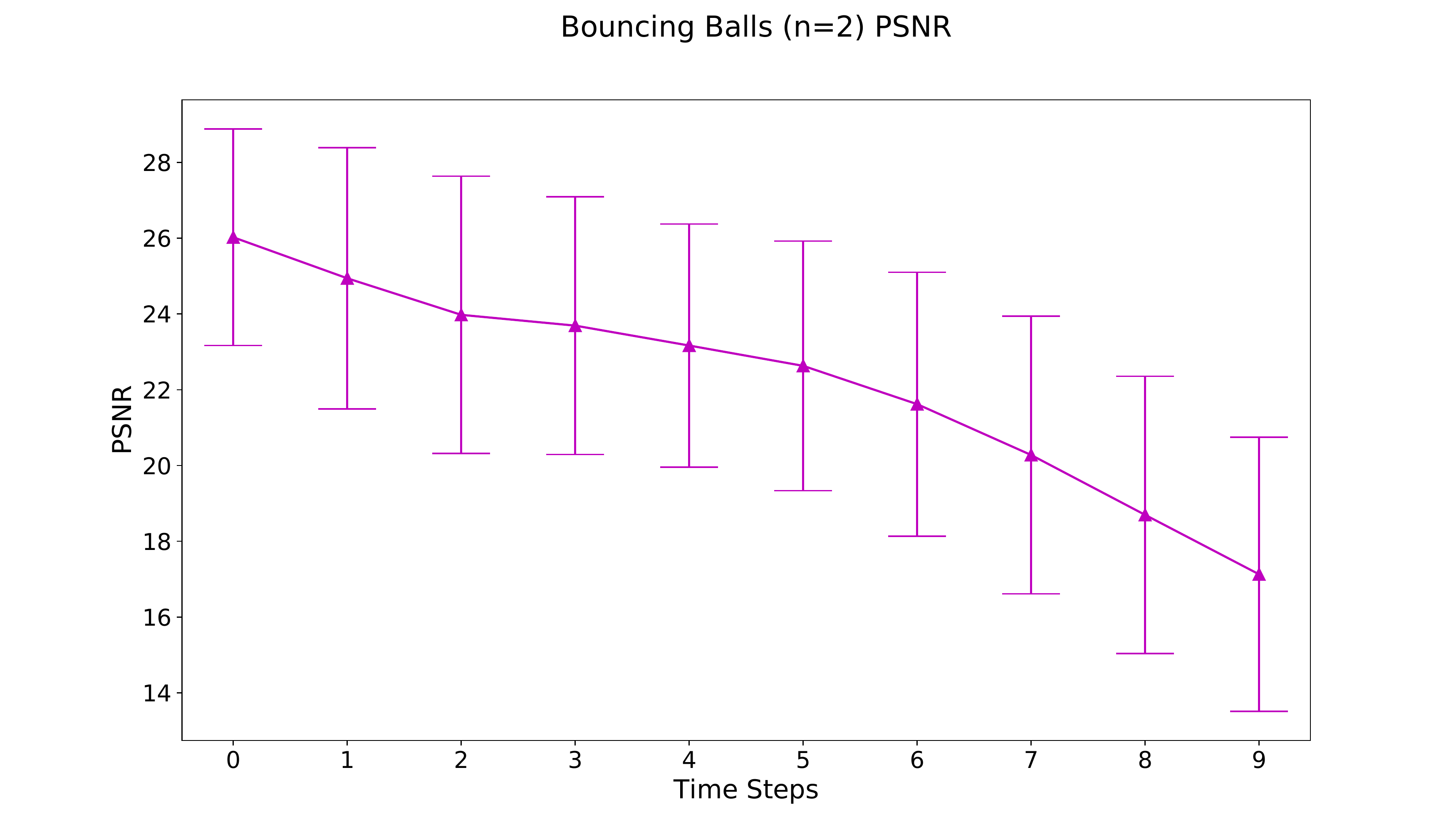}}%
\label{fig:n_2_psnr}%
\caption{MSE and PSNR values for the bouncing balls dataset with $n=2$.}
\label{fig:n_2_msepsnr}
\end{figure}

\begin{figure}[!htb]
	\begin{center}
		\includegraphics[width=0.69\columnwidth]{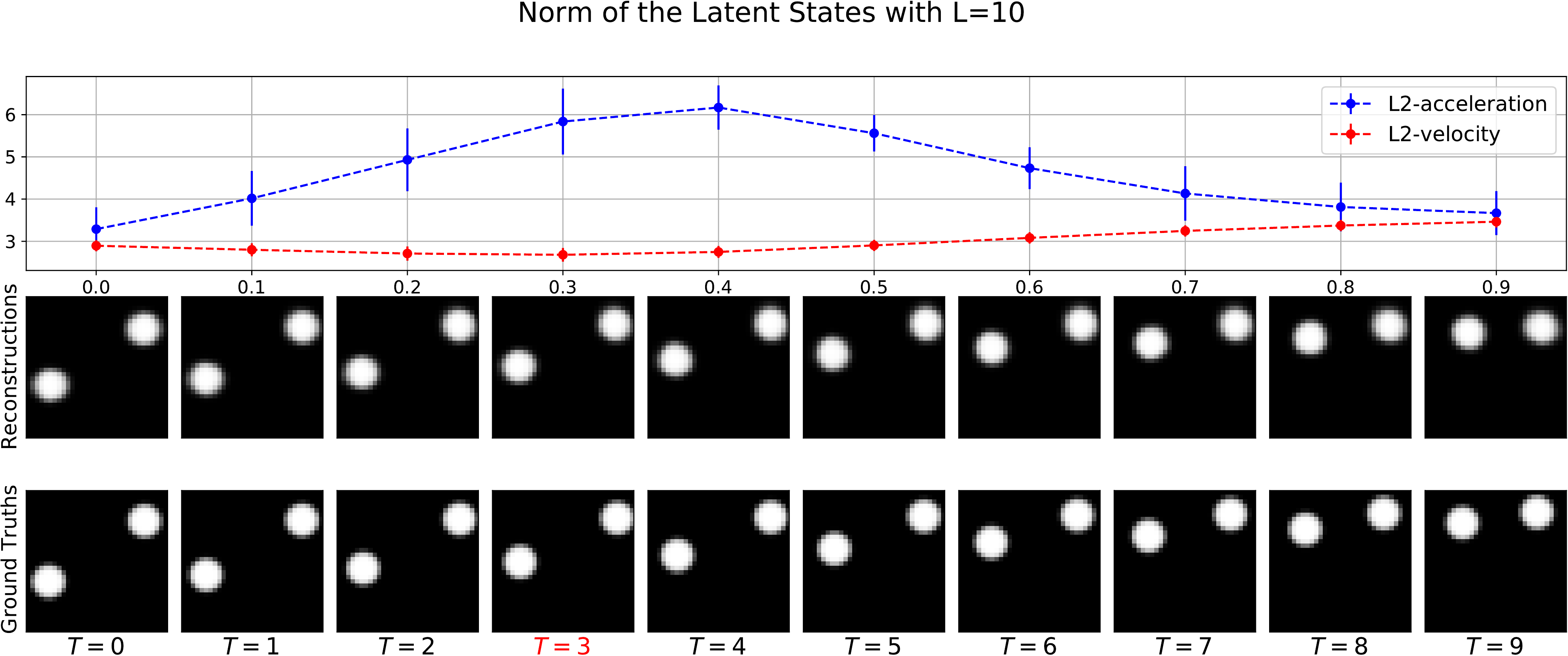}
		\caption[Example test case reconstructed by the ODE2VAE model for the bouncing ball motion with $n=2$.]{Example test case reconstructed by the ODE2VAE model with $a=5$, which is trained on the bouncing balls dataset with $n=2$. From  top  to  bottom: mean and standard deviation values of the latent norms; mean field prediction by the model with the sample size $\mathrm{L}$=10; ground truth frames. The indices of the collision times are highlighted.}
		\label{fig:n_2_figure}
	\end{center}
\end{figure}

\begin{figure}[!htb]
\centering
\subfigure[L2-norm of the Acceleration Latent]{%
\label{fig:n_2_norm_spreads}%
\includegraphics[height=1.68in]{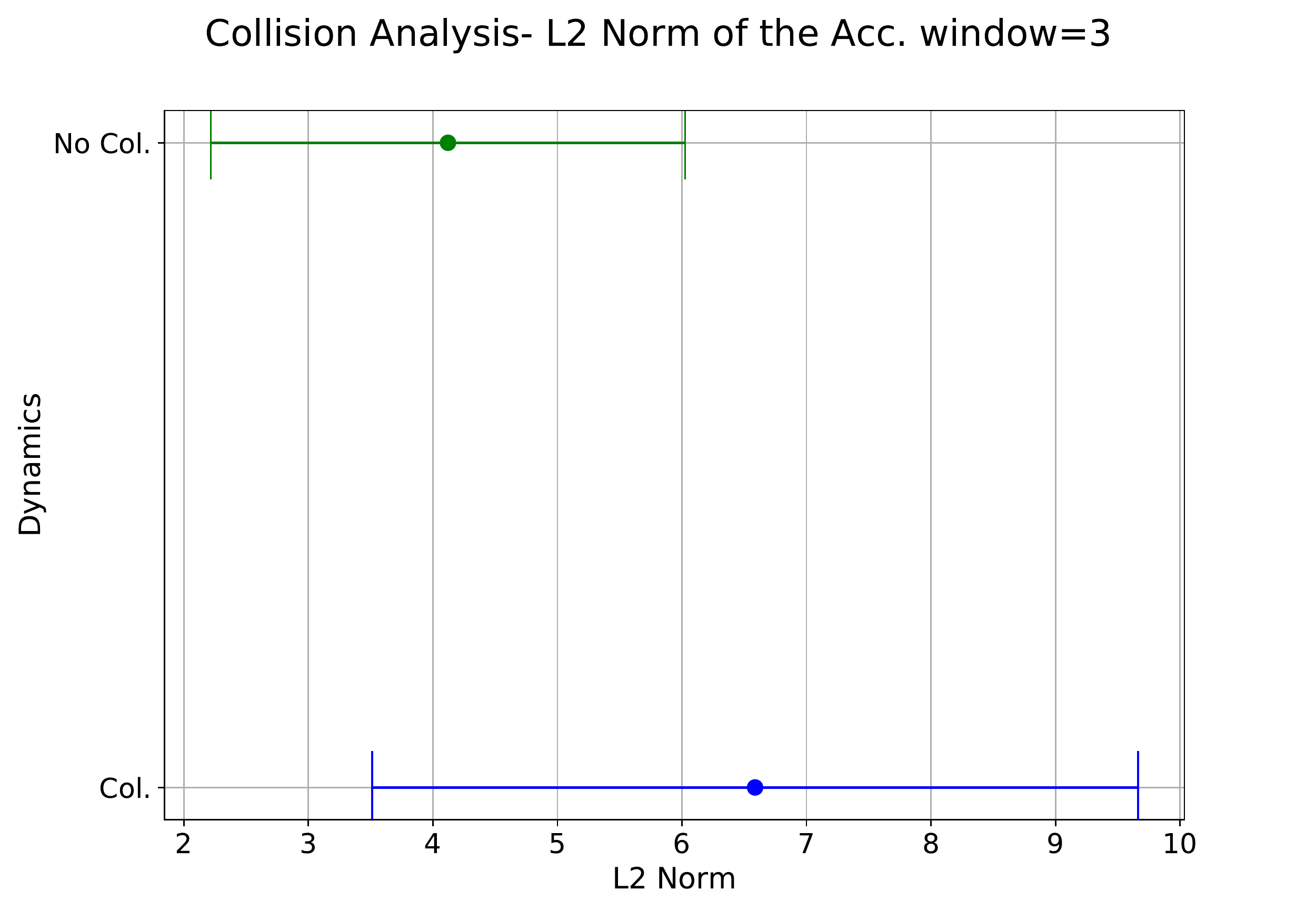}}
\qquad
\subfigure[L2-norm of the Velocity Latent]{%
\label{fig:n_2_norm_spreads2}%
\includegraphics[height=1.68in]{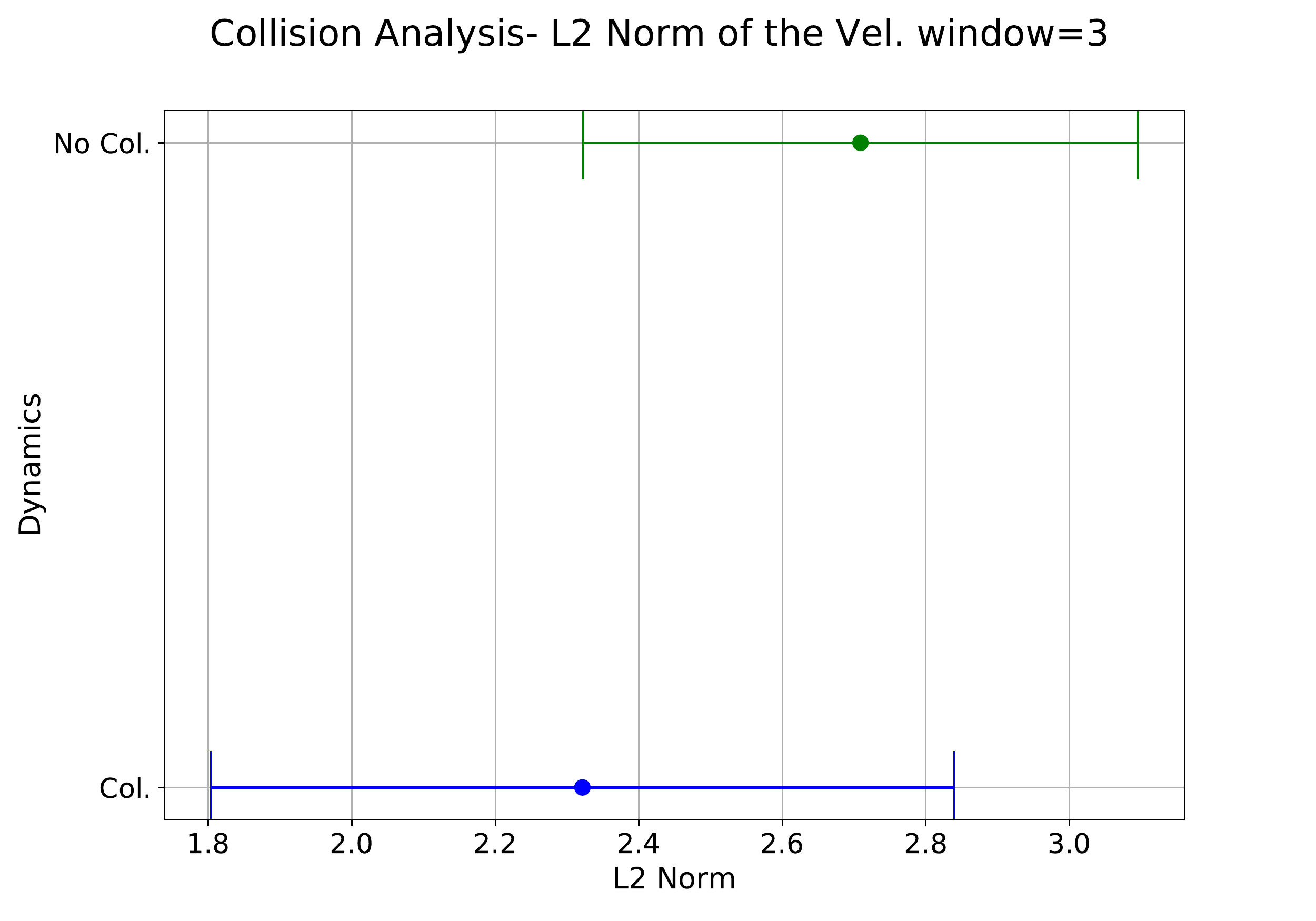}}%
\caption[Figure for the latent representation analysis of the baseline model for the bouncing balls dataset with $n=2$.]{Mean and standard deviation values for the L2-norm of the latent acceleration and velocity of the baseline model for the bouncing balls dataset with $n=2$. The figures display the values at the collision moments and the other time steps. The collision times are expanded with the window size of 3.}
\end{figure}

The baseline model with the latent dimensionality $a=5$ captures the dynamics of the motion of the two bouncing balls after it is trained for 500 epochs. Increasing the latent dimensionality to $a=6$ and $a=9$ has not increased the model performance and has caused convergence issues. In Table \ref{table:bouncing_balls_n_2}, we report the metrics for the model with $a=5$. In Figure \ref{fig:n_2_msepsnr}, we plot MSE and PSNR values for the test cases over the time steps. The model is able to capture physically meaningful latent representations to a limited extent. We present an example case in Figure \ref{fig:n_2_figure}. When the ball hits the wall, there is a spike in the norm of the latent acceleration. However, the acceleration field does not return to its initial magnitude after the collision. Although there is a slight increase in the standard deviation of the norm of the acceleration field during the collision, the uncertainty over the latent acceleration field is persistent over all the time points. Compared to the single bouncing ball case, the uncertainty of the BNN's output is less informative about the dynamics. This may stem from the increased complexity of the motion as the number of balls is doubled. Moreover, the norm of the latent velocity is not preserved during the motion. Figure \ref{fig:n_2_norm_spreads} shows that the model generates an acceleration field with a greater norm during the collisions, which is physically meaningful. However, the model is unable to generate a zero acceleration field at the time steps without collision. Figure \ref{fig:n_2_norm_spreads2} shows that the model cannot preserve the norm of the latent velocity across different test cases, which are supposed to have the same total kinetic energy in the real dynamics.


\begin{table}[!h]
\caption[Performance metrics of the selected model on the bouncing balls dataset with $n=3$.]{Performance metrics of the selected model on the bouncing balls dataset with $n=3$. For the MSE and NLL values, the lower is better. For the PSNR scores, the higher is better. Each metric is computed by using 10 samples per test case.}
\centering
\begin{tabular}{c *{3}{S[table-format=2,
                         table-column-width=7em]}
                }
    \toprule
            & \multicolumn{3}{c}{Metrics} 
                         \\
    \cmidrule(lr){2-4}
Model  & {MSE}    & {PSNR}    & {NLL}    \\
    \midrule
ODE2VAE, $a=8$&{$0.0196\pm0.0174$}&{$18.4438\pm3.4380$}&{$154.9332$}\\
    \bottomrule
\end{tabular}
\label{table:bouncing_balls_n_3}

    \end{table}

\begin{figure}[!htb]
\centering
\subfigure[MSE values for the bouncing balls dataset with $n=3$.]{%
\includegraphics[height=1.73in]{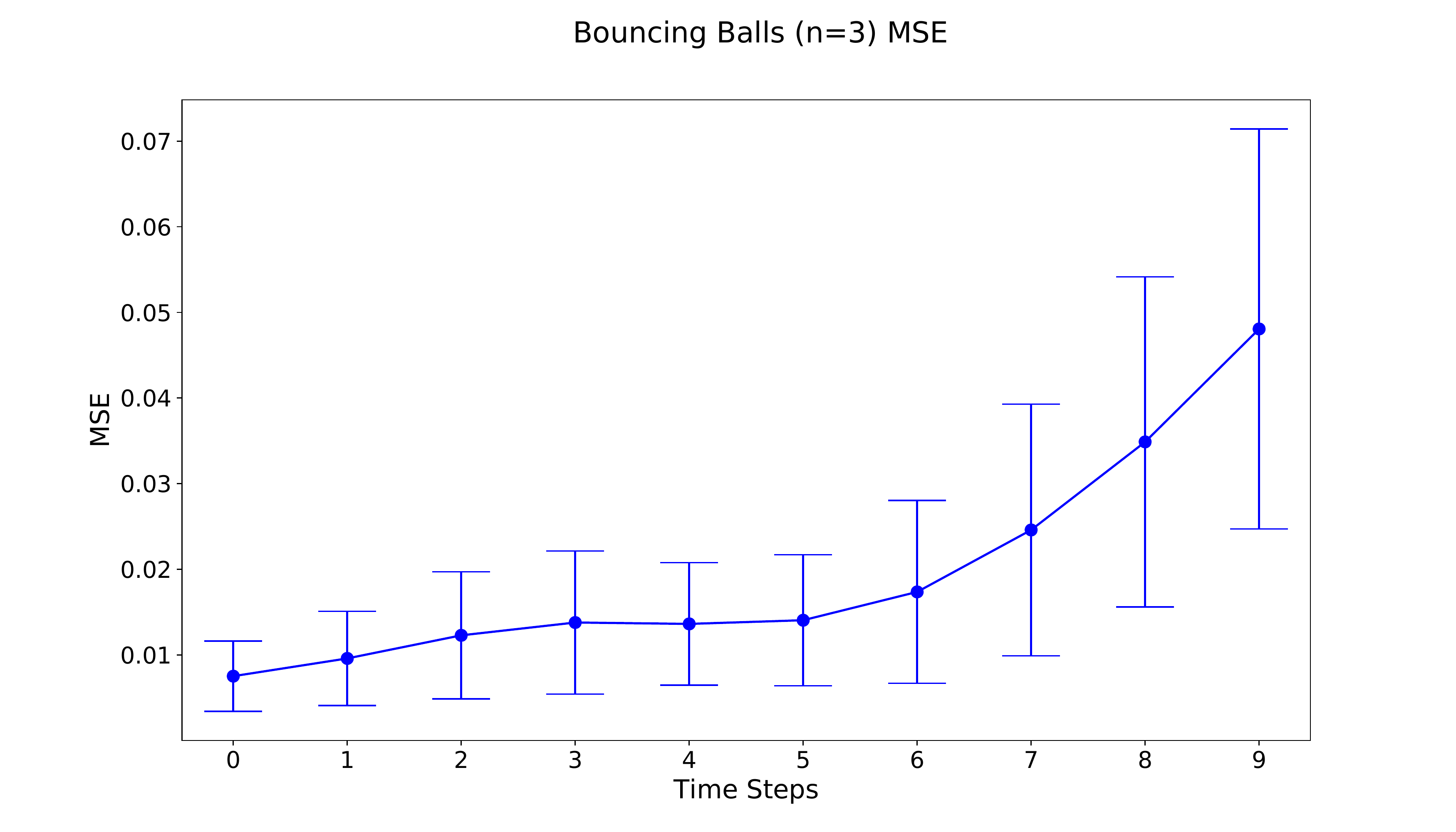}}
\label{fig:n_3_mse}
\qquad
\subfigure[PSNR values for the bouncing balls dataset with $n=3$.]{%
\includegraphics[height=1.73in]{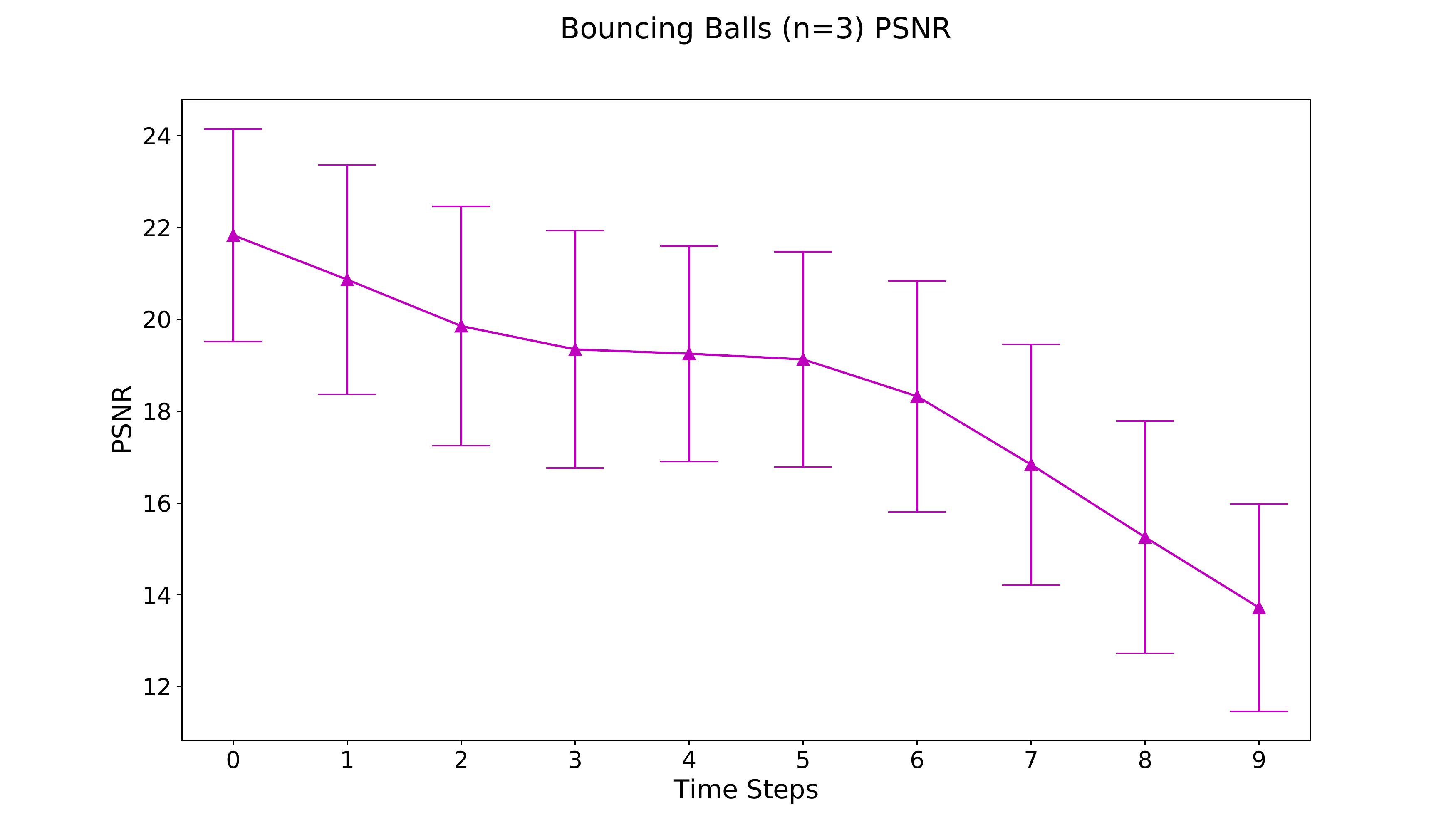}}%
\label{fig:n_3_psnr}%
\caption{MSE and PSNR values for the bouncing balls dataset with $n=3$.}
\label{fig:n_3_msepsnr}
\end{figure}

\begin{figure}[!htb]
	\begin{center}
		\includegraphics[width=0.69\columnwidth]{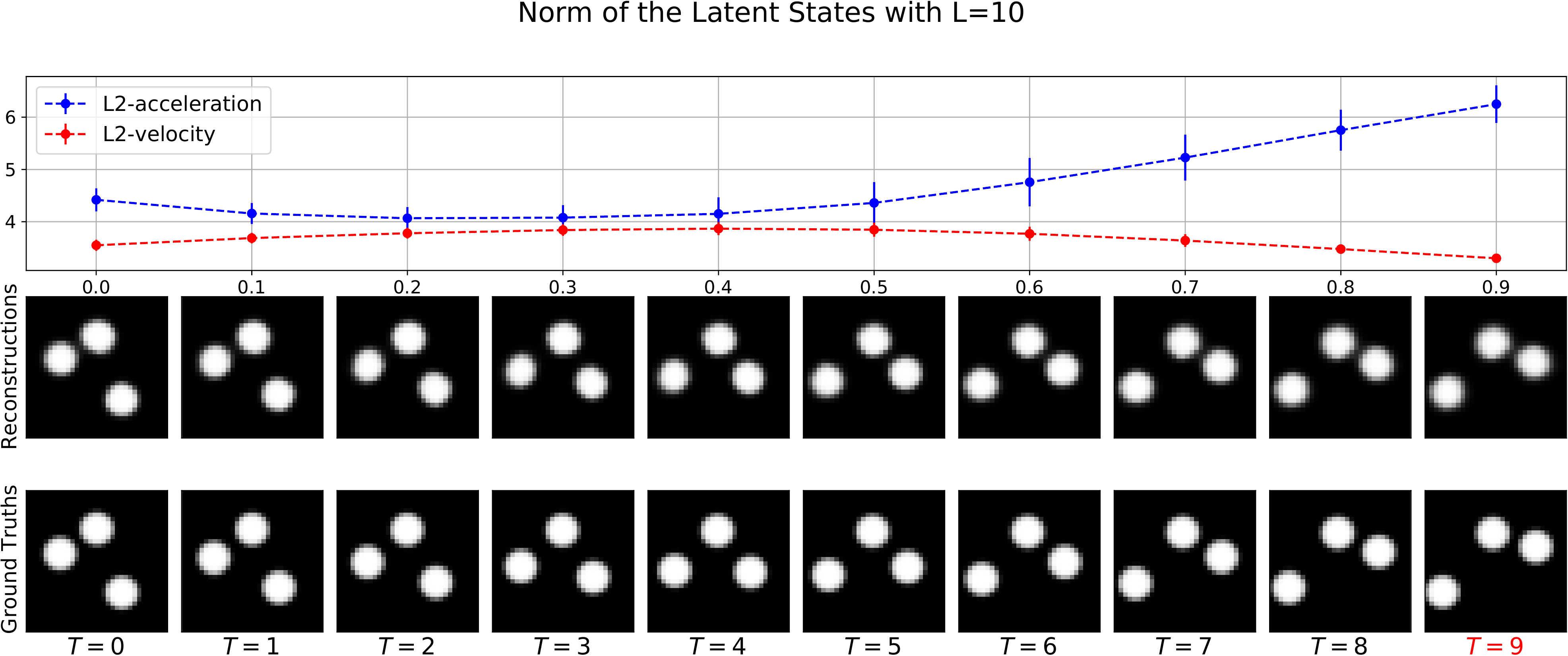}
		\caption[Example test case reconstructed by the ODE2VAE model for the bouncing ball motion with $n=3$.]{Example test case reconstructed by the ODE2VAE model with $a=8$, which is trained on the bouncing balls dataset with $n=3$. The first row of the figure shows how the dynamics of the latent norms evolve, the second row shows the model's mean field predictions, and the last row shows ground truth frames. The indices of the collision times are highlighted.}
		\label{fig:n_3_figure}
	\end{center}
\end{figure}

\begin{figure}[!h]
\centering
\subfigure[L2-norm of the Acceleration Latent]{%
\label{fig:n_3_norm_spreads}%
\includegraphics[height=1.68in]{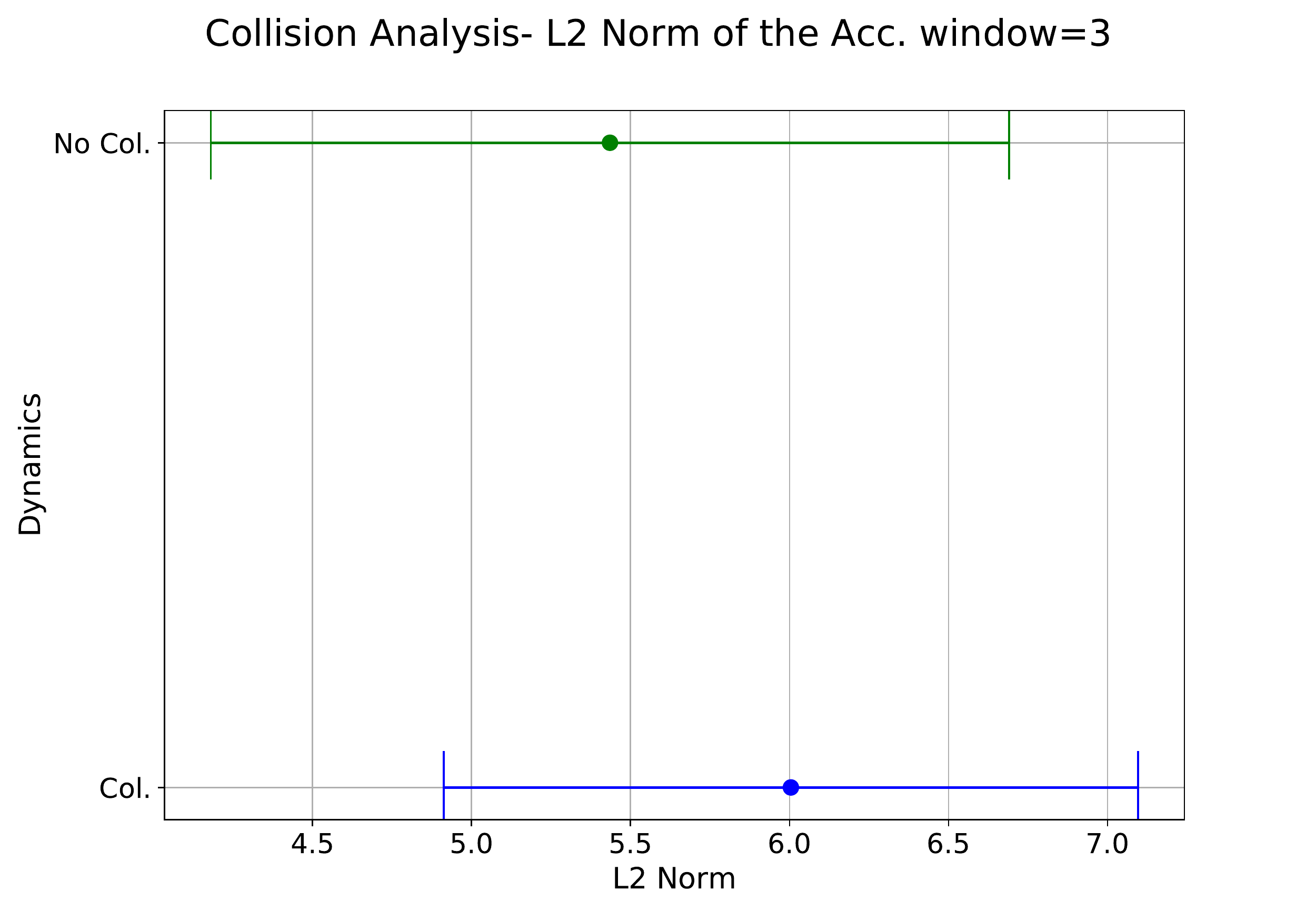}}
\qquad
\subfigure[L2-norm of the Velocity Latent]{%
\label{fig:n_3_norm_spreads2}%
\includegraphics[height=1.68in]{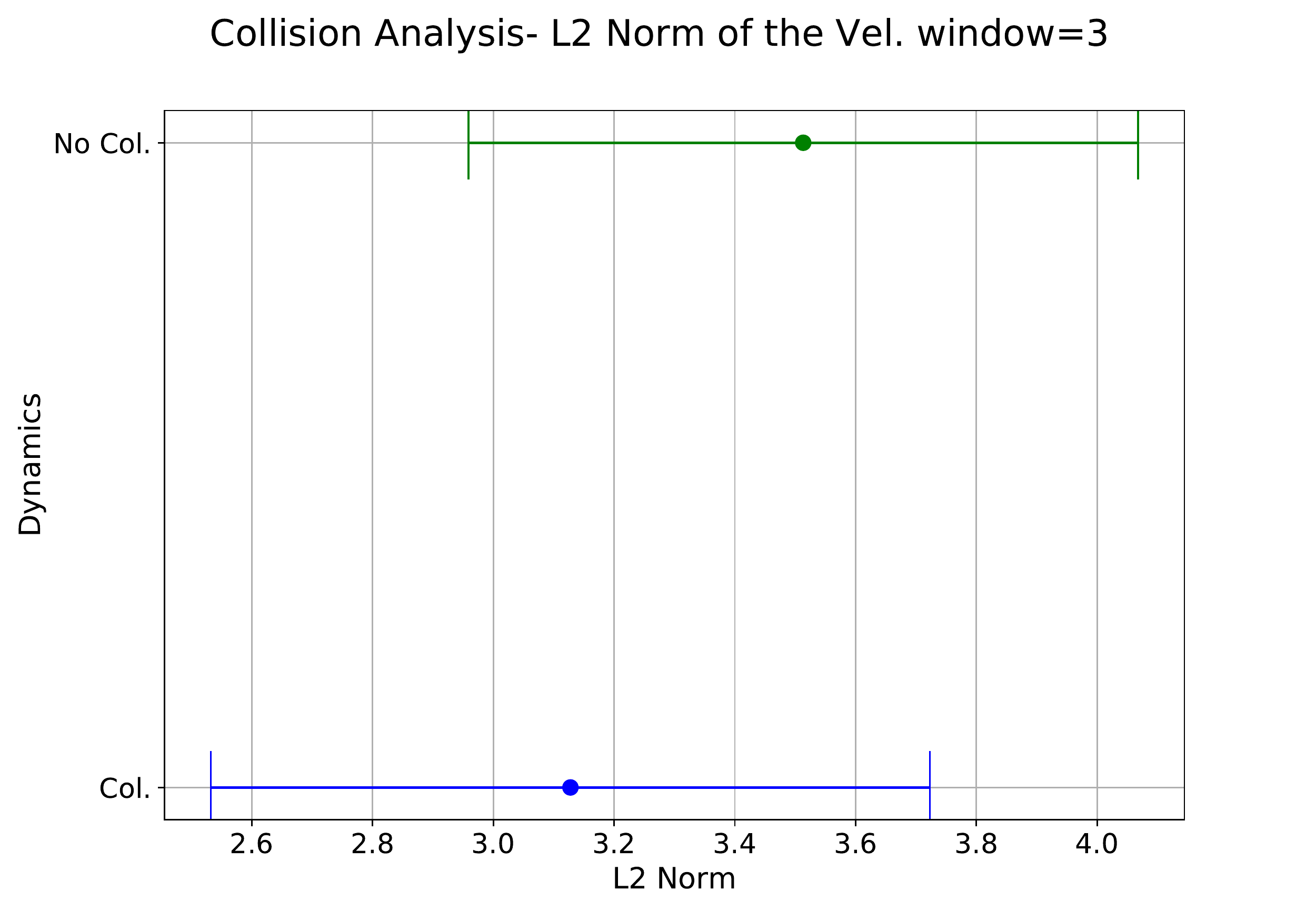}}%
\caption[Figure for the latent representation analysis of the baseline model for the bouncing balls dataset with $n=3$.]{Mean and standard deviation values for the L2-norm of the latent acceleration and velocity of the baseline model for the bouncing balls dataset with $n=3$. The figures display the values at the collision moments and the other time steps. The collision times are expanded with the window size of 3.}
\end{figure}

We find out that the ODE2VAE model could not capture the motion of three bouncing balls with $a=7$. The baseline model with the latent dimensionality $a=8$ captures the motion of the three bouncing balls after it is trained for 1000 epochs. Increasing the latent dimensionality to $a=12$ has not increased the model performance. In Table \ref{table:bouncing_balls_n_3}, we report the metrics for the model with $a=8$. In Figure \ref{fig:n_3_msepsnr}, we plot MSE and PSNR values for the test cases over the time steps. As expected there is an increasing trend in MSE scores and decreasing trend in PSNR scores. We present an example case in Figure \ref{fig:n_3_figure}. A careful observation of the reconstructions for the time steps three and four indicates that the model fails to preserve a rigid ball shape. The last frame of the reconstructions also shows that the model lags behind and misses the ball to wall collision. We observe that the magnitude of the acceleration field increases during the collision. However, it starts increasing prior to the collision moments. The uncertainty over the acceleration field's magnitude increases as the model extrapolates into the future. There may be two reasons behind this increased uncertainty, which are the model's reduced predictive performance during the extrapolations and the non-linear motion during the collision. Figure \ref{fig:n_3_norm_spreads} shows that the model still generates an acceleration field with a higher magnitude for the time steps with collision. We observe that as the real dynamics get more complex, the model tends to create acceleration fields with closer magnitudes for the collision and no collision moments. Similar to the previous motions, the model is unable to preserve the norm of latent velocity across different test cases, which is shown in Figure \ref{fig:n_3_norm_spreads2}.

\vspace*{0.5\baselineskip}
\vspace*{1\baselineskip}
\subsection{Simple Pendulum}
\begin{figure}[!htb]
	\begin{center}
		\includegraphics[width=0.5\columnwidth]{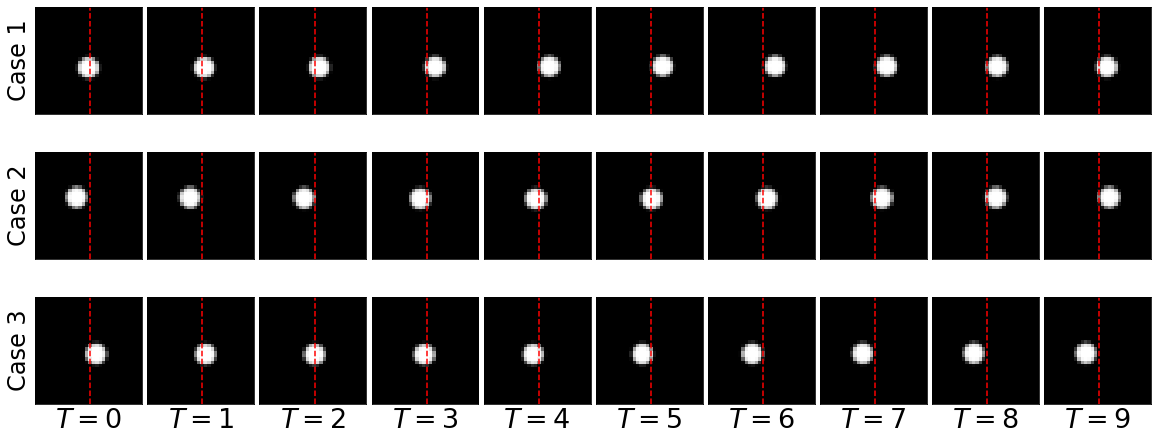}
		\caption[Figure for the simple pendulum dataset.]{Simple pendulum dataset with single point mass hanged from the pivoted rod. Since the aim is to learn the dynamics of the ball, the rod is not visualized in the frames. The vertical lines denote the axis that passes through the equilibrium point. The sequence length T is 10.}
		\label{fig:pendulum}
	\end{center}
\end{figure}

A simple pendulum system consists of a point mass, and a pivoted rod with length $l$, where the mass is suspended from the rod. The mass of the rod is negligible, and there is no air friction in our case. In this paper, we only consider pendulum motions with small initial angles. Therefore, it is sufficient to model simple pendulum motion. The dynamics of the simple pendulum is approximated by using small angle approximation, $\sin{\alpha} \approx \alpha$. The dataset captures the periodic motion of the point mass around its equilibrium position. During the motion, the gravitational field is uniform. The object reaches its maximum kinetic energy at its equilibrium position. The magnitude of the restoring force over the object increases as it approaches its highest point of swing. The side length of the 2D square box is $\SI{10.0}{\meter}$. The radius of point mass is $\SI{1.0}{\meter}$. The length of the rod $l$ has a uniform distribution in the range $[3,6]$. The initial angle for freeing the point mass follows a uniform distribution in the range $[\pi/36, \pi/9]$. The gravitational field has a magnitude of $\SI{9.91}{\meter/\second^2}$. The frames are separated by 0.4 seconds, and the motion is simulated by the analytical solution for the simple pendulum motion. Due to the different initial angles, the maximum total kinetic energy for each case is different. Figure \ref{fig:pendulum} shows example sequences from the pendulum dataset.

    \begin{table}[thbp]
\caption[Performance metrics of the selected model on the simple pendulum dataset.]{Performance metrics of the selected model on the simple pendulum dataset. Each metric is computed by using 10 samples per test case.}
\centering
\begin{tabular}{c *{3}{S[table-format=2,
                         table-column-width=7em]}
                }
    \toprule
            & \multicolumn{3}{c}{Metrics} 
                         \\
    \cmidrule(lr){2-4}
Model  & {MSE}    & {PSNR}    & {NLL}    \\
    \midrule
ODE2VAE, $a=2$&{$0.0007\pm0.0006$}&{$33.6325\pm4.5241$}&{$26.5609$}\\
    \bottomrule
\end{tabular}
\label{table:simple_pendulum}
    \end{table}

\begin{figure}[!htb]
\centering
\subfigure[MSE values for the pendulum dataset.]{%
\includegraphics[height=1.73in]{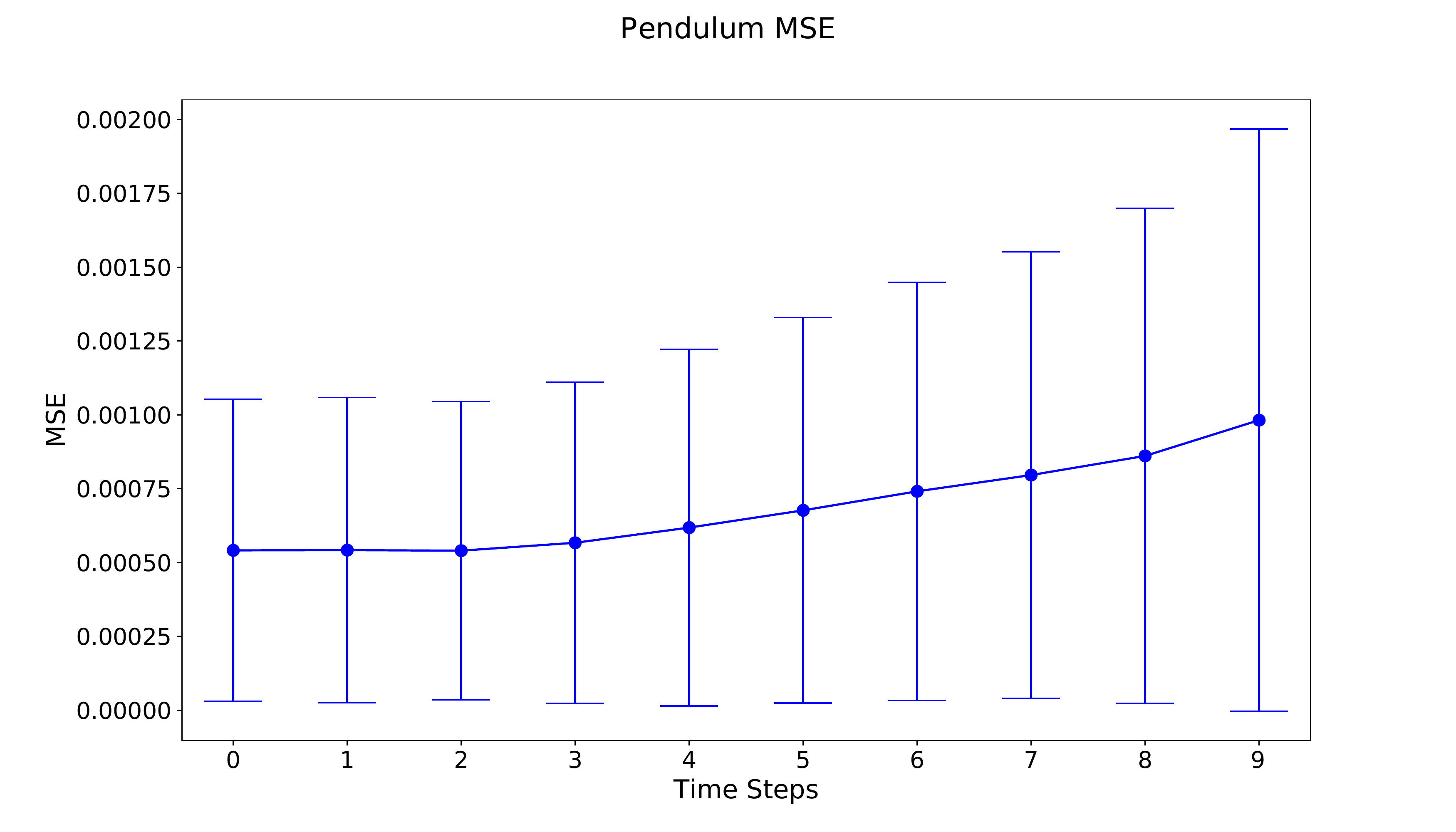}}
\label{fig:pendulum_mse}
\qquad
\subfigure[PSNR values for the pendulum dataset.]{%
\includegraphics[height=1.73in]{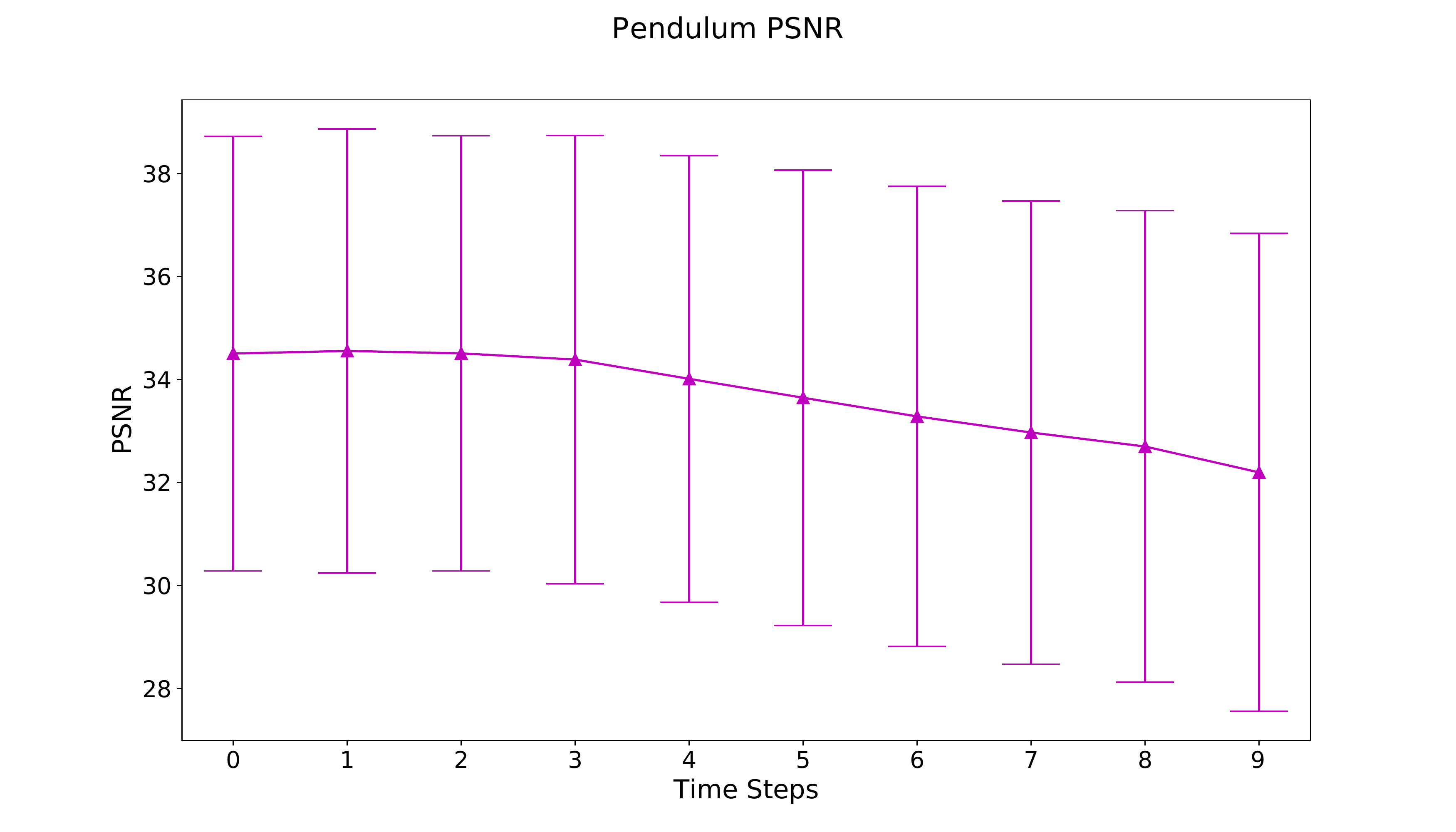}}%
\label{fig:pendulum_psnr}%
\caption{MSE and PSNR values for the pendulum dataset.}
\label{fig:pendulum_mse_psnr}
\end{figure}
\begin{figure}[!htb]
	\begin{center}
		\includegraphics[width=0.69\columnwidth]{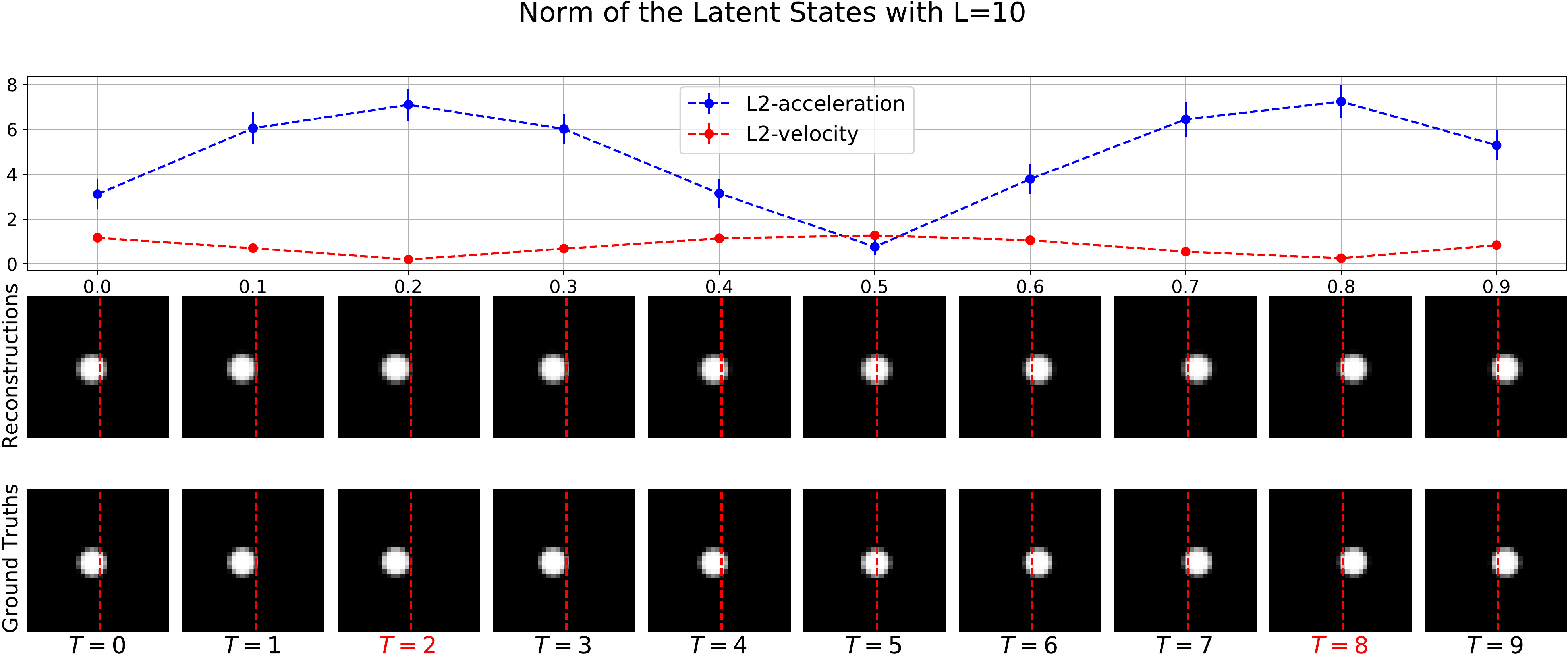}
				\caption[Example test case reconstructed by the ODE2VAE model for the simple pendulum motion.]{Example test case reconstructed by the ODE2VAE model with $a=2$, which is trained on the simple pendulum dataset. From  top  to  bottom: mean and standard deviation values of the latent norms; mean field prediction by the model with the sample size $\mathrm{L}$=10; ground truth frames. The indices with a direction change are highlighted. The vertical lines on the images denote the equilibrium axis of the pendulum.}
				
		\label{fig:pendulum_figure}
	\end{center}
\end{figure}
\begin{figure}[!h]
\centering
\subfigure[L2-norm of the Acceleration Latent]{%
\label{fig:pendulum_norm_spread_acc}%
\includegraphics[height=1.68in]{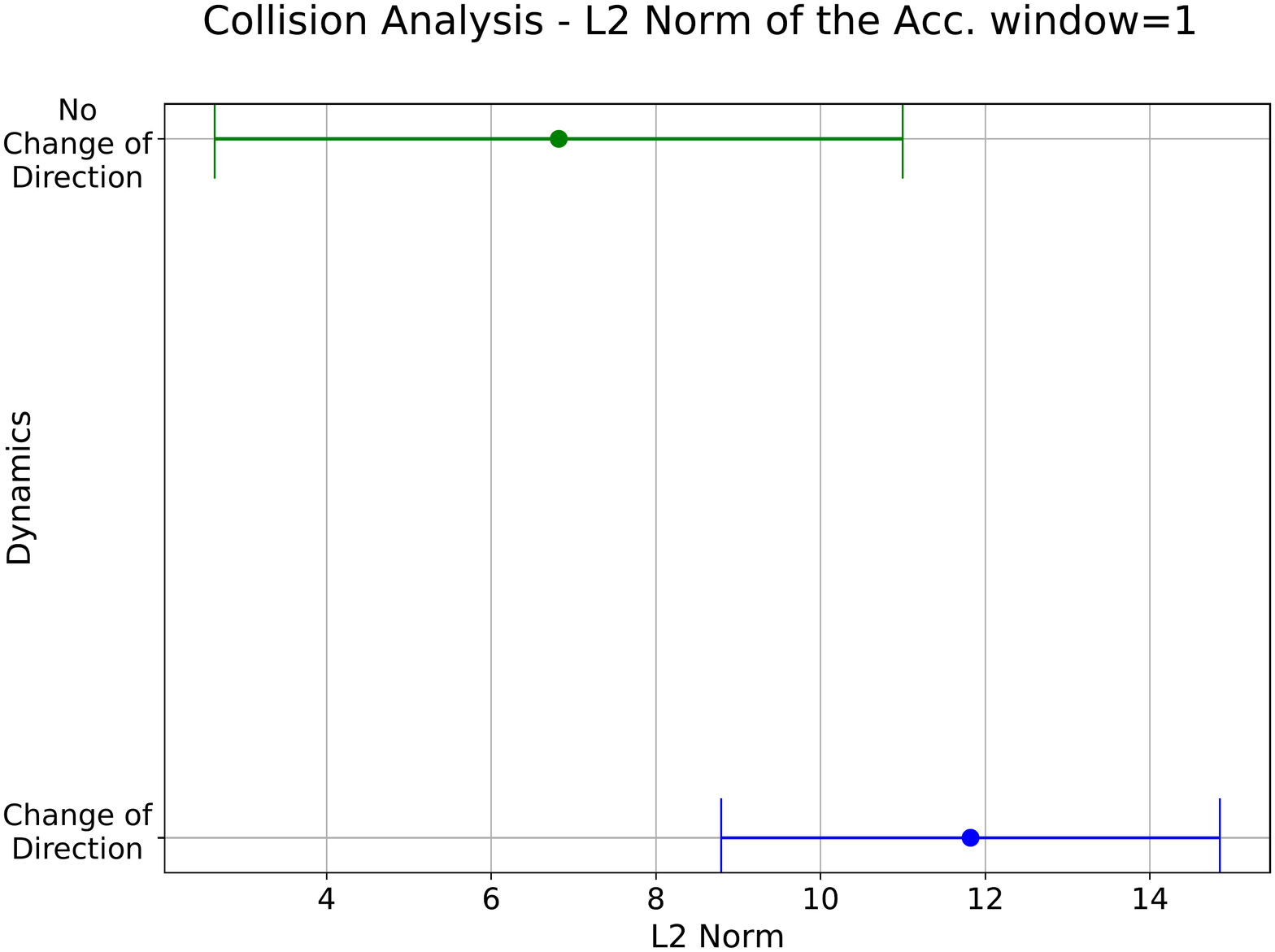}}%
\qquad
\subfigure[L2-norm of the Velocity Latent]{%
\label{fig:pendulum_norm_spread_vel}%
\includegraphics[height=1.68in]{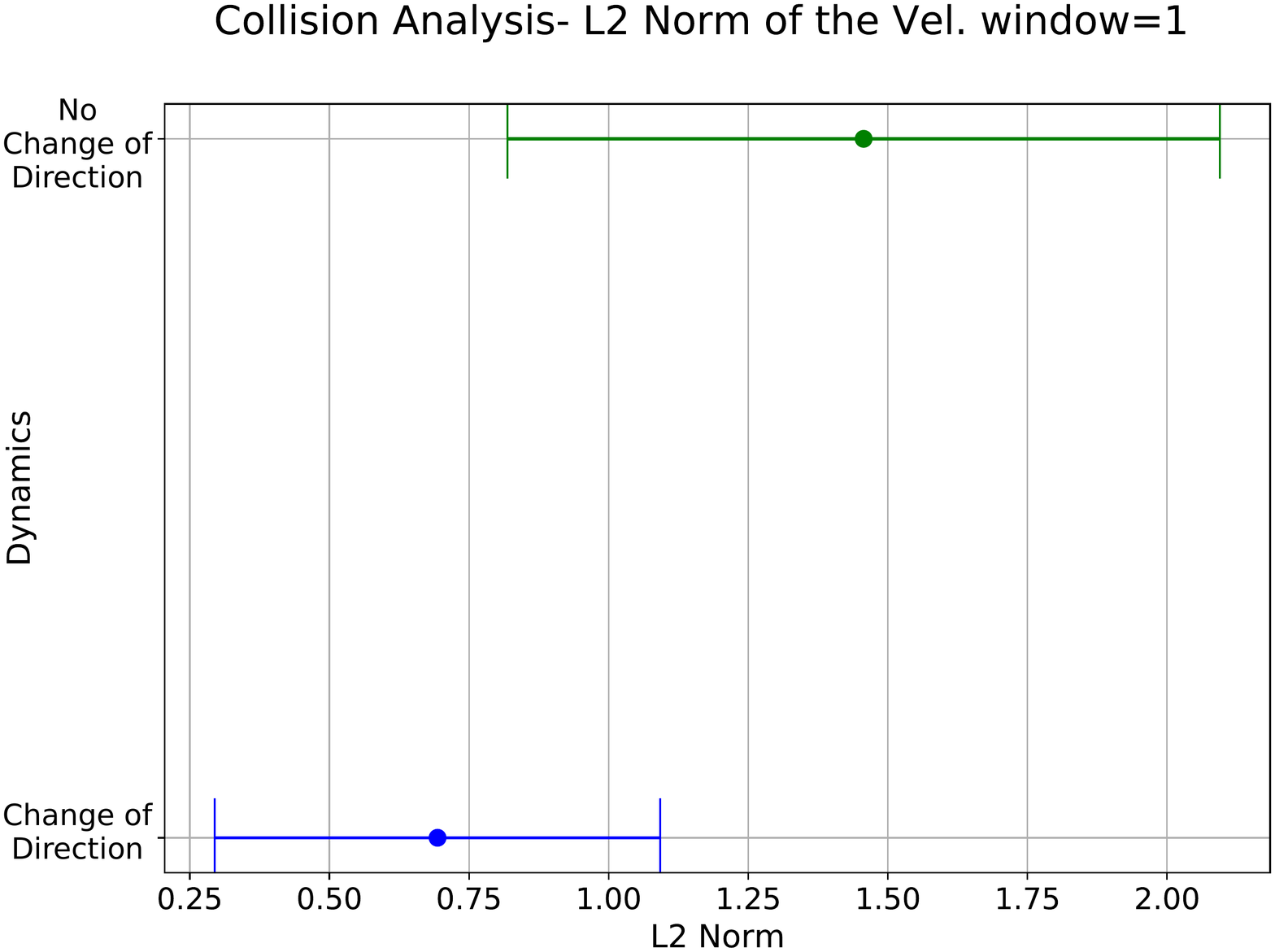}}%
\caption[Figure for the latent representation analysis of the baseline model for the pendulum dataset.]{Mean and standard deviation values for the L2-norm of the latent acceleration and velocity the pendulum dataset. The figures display the values at the moment of direction change and other time steps.}
\end{figure}

We train the ODE2VAE model with $a=2$ and $a=6$ on the simple pendulum dataset for 300 epochs. Both models have captured the pendulum motion; there is not a considerable improvement between the two models. Since the true pendulum motion can be modeled with two dimensions, the only way for the model to capture the motion with two latent dimensions is by using the decoder to capture the radius of the ball. In Table \ref{table:simple_pendulum}, we only report the model's performance with $a=2$. In Figure \ref{fig:pendulum_mse_psnr}, we plot MSE and PSNR values for the test cases over the time steps. Since the simple pendulum motion is a periodic motion, it may be easier to capture compared to the other motion types. The model is able to capture physically meaningful latent representations. We provide an example case in Figure \ref{fig:pendulum_figure}. The model generates an increased norm of the latent acceleration when the object reaches its highest point during its motion and the magnitude of the acceleration is minimized when the object passes through the equilibrium point. Another observation is that the BNN has the decreased uncertainty over the magnitude of the acceleration field when the object passes through the equilibrium point. Additionally, the norm of the latent velocity reaches its peak when the ball passes the equilibrium position and decreases when the object reaches its highest points, which are the highlighted time points. Although the model has the latent dimensionality $a=2$, the latent dynamics of the model behave similarly to the true dynamics of the simple pendulum. In Figure \ref{fig:pendulum_norm_spread_acc}, we provide the statistics for the norm of the latent acceleration over the test cases with a breakdown for the moments with and without direction change. The figure shows that the model generates an acceleration field with a greater magnitude when the object reaches its highest point. Moreover, Figure \ref{fig:pendulum_norm_spread_vel} shows that the latent velocities of the test cases have a reduced norm at the moments of direction change. Although the model cannot generate latent velocity variables with magnitude zero at the moment of direction change, it captures a decreasing trend, which is still physically plausible.


\subsection{Projectile Motion}
\begin{figure}[!htb]
	\begin{center}
		\includegraphics[width=0.5\columnwidth]{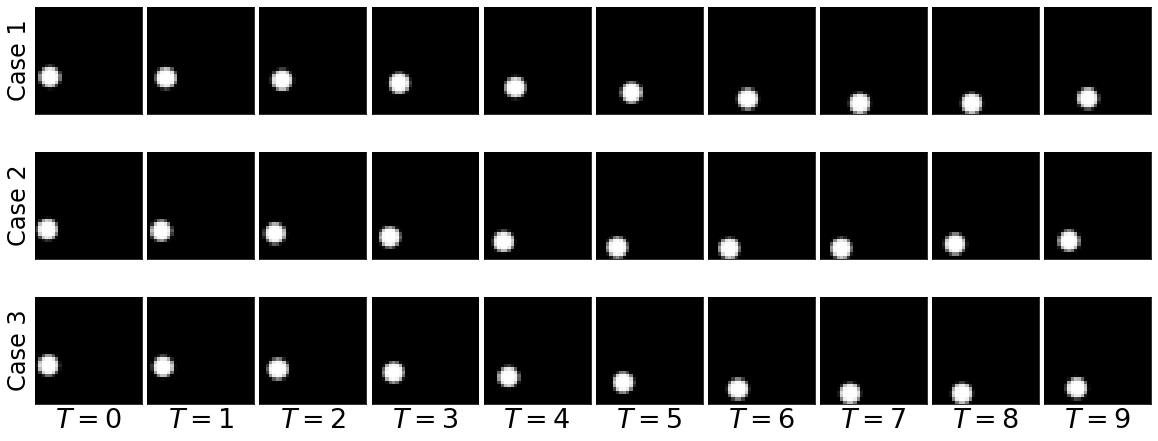}
		\caption[Figure for the projectile motion dataset.]{Projectile motion dataset with a ball launched from a initial height. The sequence length T is 10.}
		\label{fig:projectile}
	\end{center}
\end{figure}

The projectile motion dataset consists of a ball, which is projected up in a square frame with the side length of $\SI{10.0}{\meter}$. The ball is affected by gravity and collisions during its motion. The ball reaches its maximum kinetic energy before its first collision with the ground. During the collision it loses some of its kinetic energy. It has a radius of $\SI{1.0}{\meter}$. The initial velocity ($\SI{}{\meter/\second}$) is denoted as $v=[v_{x},v_{y}]$, where $v_{x}$ and $v_{y}$ are sampled from a uniform distribution with the ranges $[1,4]$ and $[0,1]$. The initial position ($\SI{}{\meter}$) is denoted as $h=[h_{x},h_{y}]$, where $h_{x}$ is fixed as zero and $h_{y}$ has a uniform distribution in the range $[1,3]$. The coefficient of restitution, which determines the magnitude of the $v_{y}$ after the ball hits and bounces from the floor, is $0.80$. The collision between the ball and floor is assumed to take $0.1$ second. The frames are separated by 0.1 second. Figure \ref{fig:projectile} shows example sequences from the projectile motion dataset.

    \begin{table}[thbp]
\caption[Performance metrics of the selected model on the projectile motion dataset.]{Performance metrics of the selected model on the projectile motion dataset. Each metric is computed by using 10 samples per test case.}
\centering
\begin{tabular}{c *{3}{S[table-format=2,
                         table-column-width=7em]}
                }
    \toprule
            & \multicolumn{3}{c}{Metrics} 
                         \\
    \cmidrule(lr){2-4}
Model  & {MSE}    & {PSNR}    & {NLL}    \\
    \midrule
    ODE2VAE, $a=9$&{$0.0016\pm0.0016$}&{$30.5322\pm5.1913$}&{$27.8267$}
    \\
    \bottomrule
\end{tabular}
\label{table:projectile}
    \end{table}

\begin{figure}[!htb]
\centering
\subfigure[MSE values for the projectile motion dataset.]{%
\includegraphics[height=1.73in]{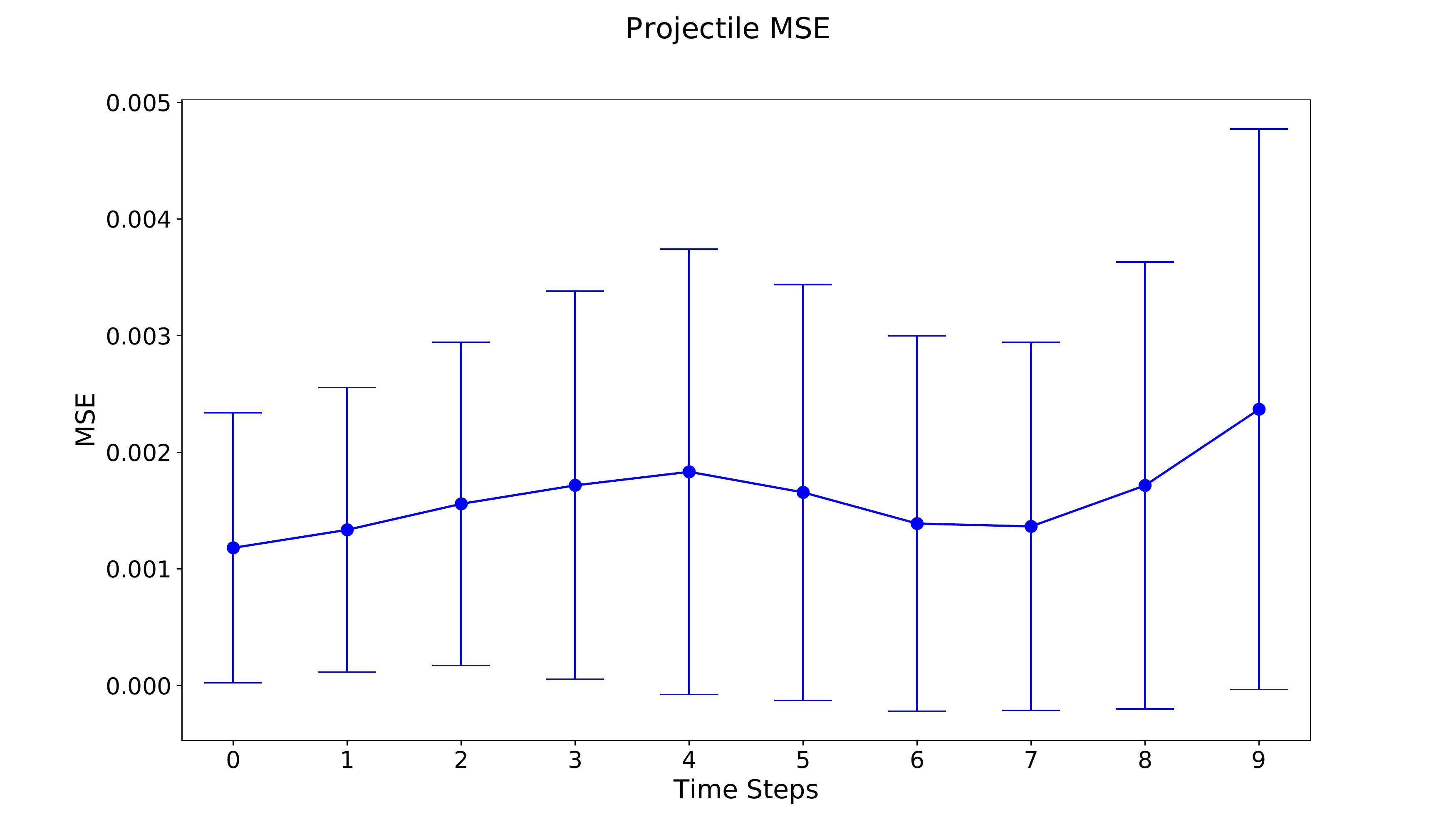}}
\label{fig:projectile_mse}
\qquad
\subfigure[PSNR values for the projectile motion dataset]{%
\includegraphics[height=1.73in]{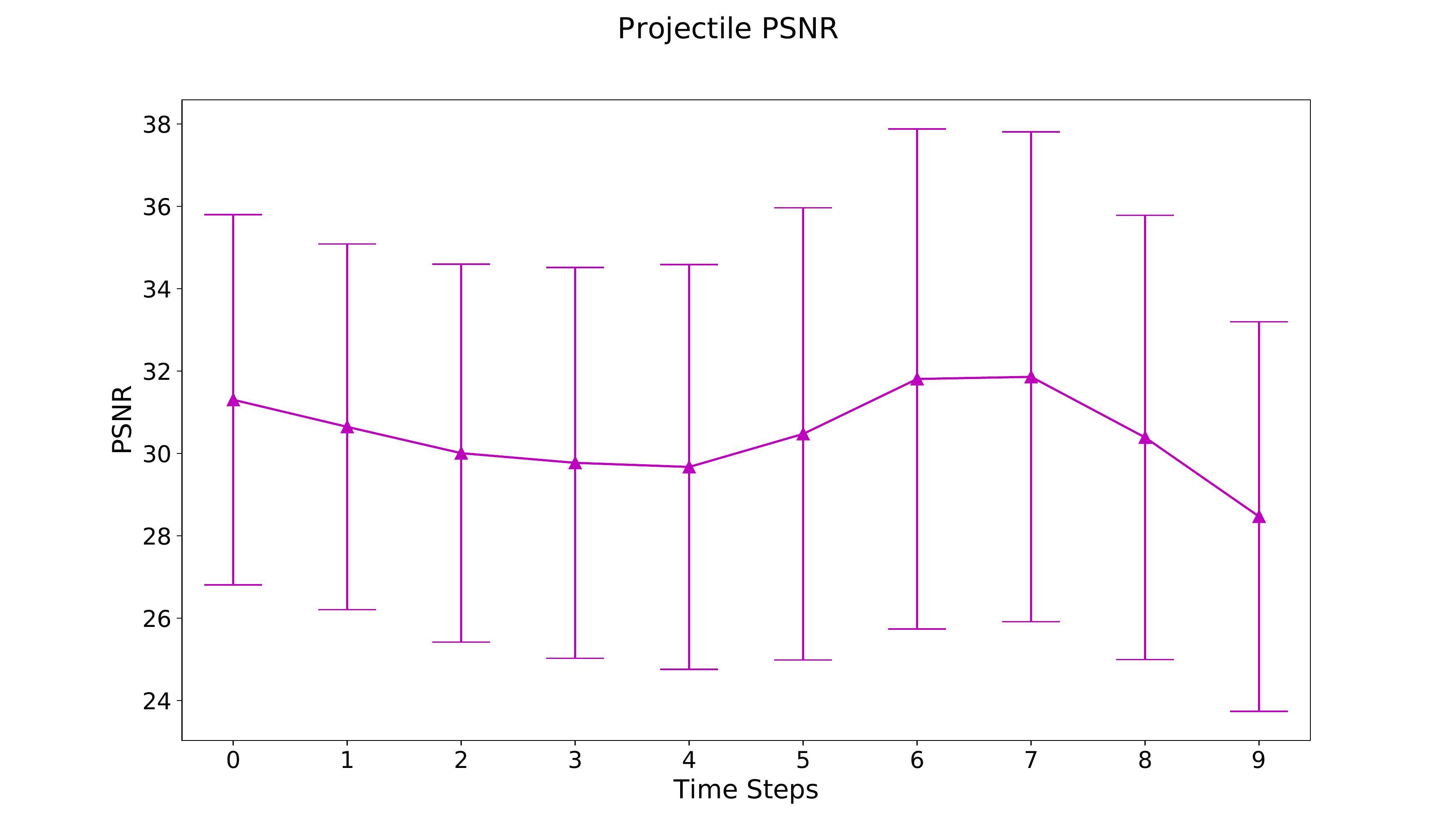}}%
\label{fig:projectile_psnr}%
\caption{MSE and PSNR values for the projectile motion dataset.}
\label{fig:projectile_msepsnr}
\end{figure}

\begin{figure}[!htb]
	\begin{center}
		\includegraphics[width=0.69\columnwidth]{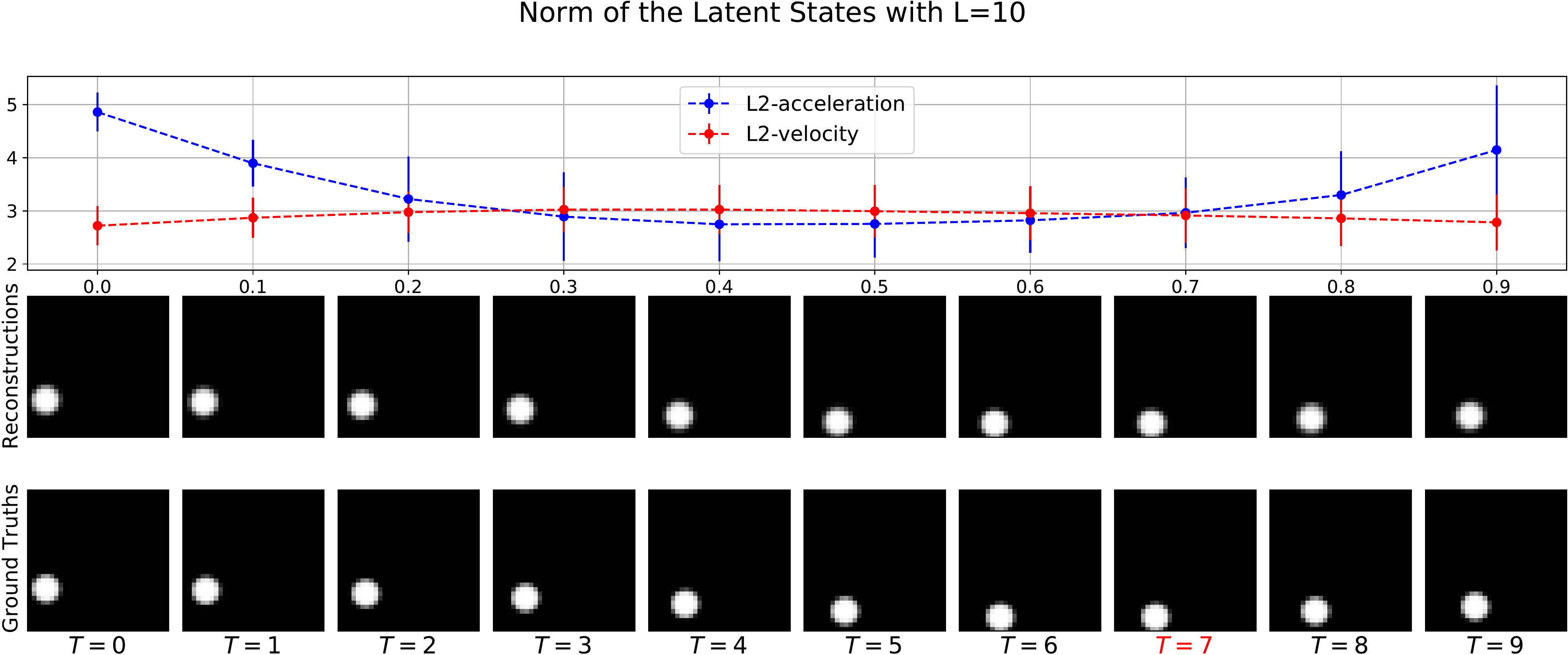}
			\caption[Example test case reconstructed by the ODE2VAE model for the simple pendulum motion.]{Example test case reconstructed by the ODE2VAE model with $a=9$, which is trained on the projectile motion dataset. From  top  to  bottom: mean and standard deviation values of the latent norms; mean field prediction by the model with the sample size $\mathrm{L}$=10; ground truth frames. The indices with a collision are highlighted.}
		\label{fig:projectile_figure}
	\end{center}
\end{figure}

\begin{figure}[!htb]
\centering
\subfigure[L2-norm of the Acceleration Latent]{%
\label{fig:projectile_norm_spreads_acc}%
\includegraphics[height=1.68in]{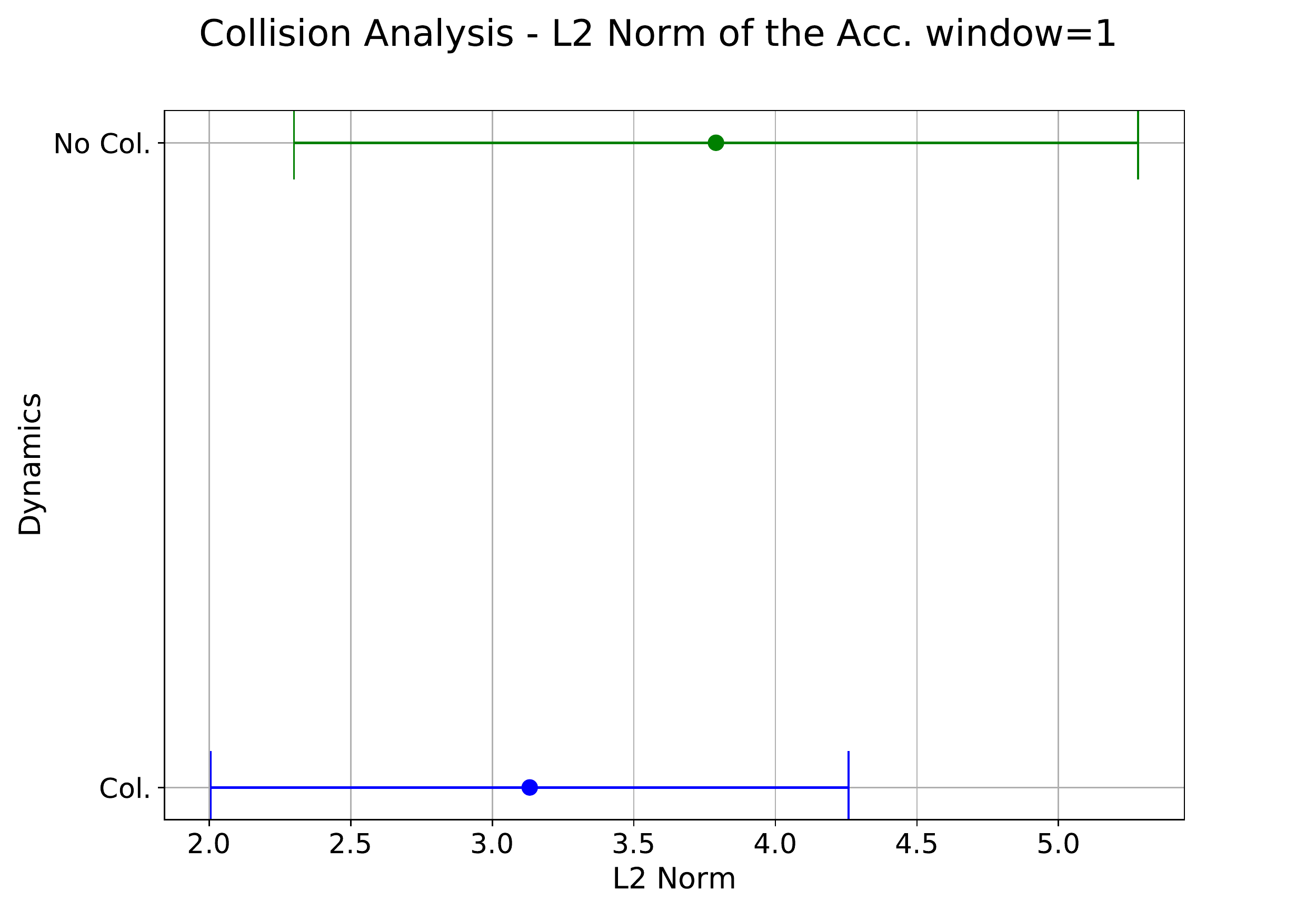}}%
\qquad
\subfigure[L2-norm of the Velocity Latent]{%
\label{fig:projectile_norm_spreads_vel}%
\includegraphics[height=1.68in]{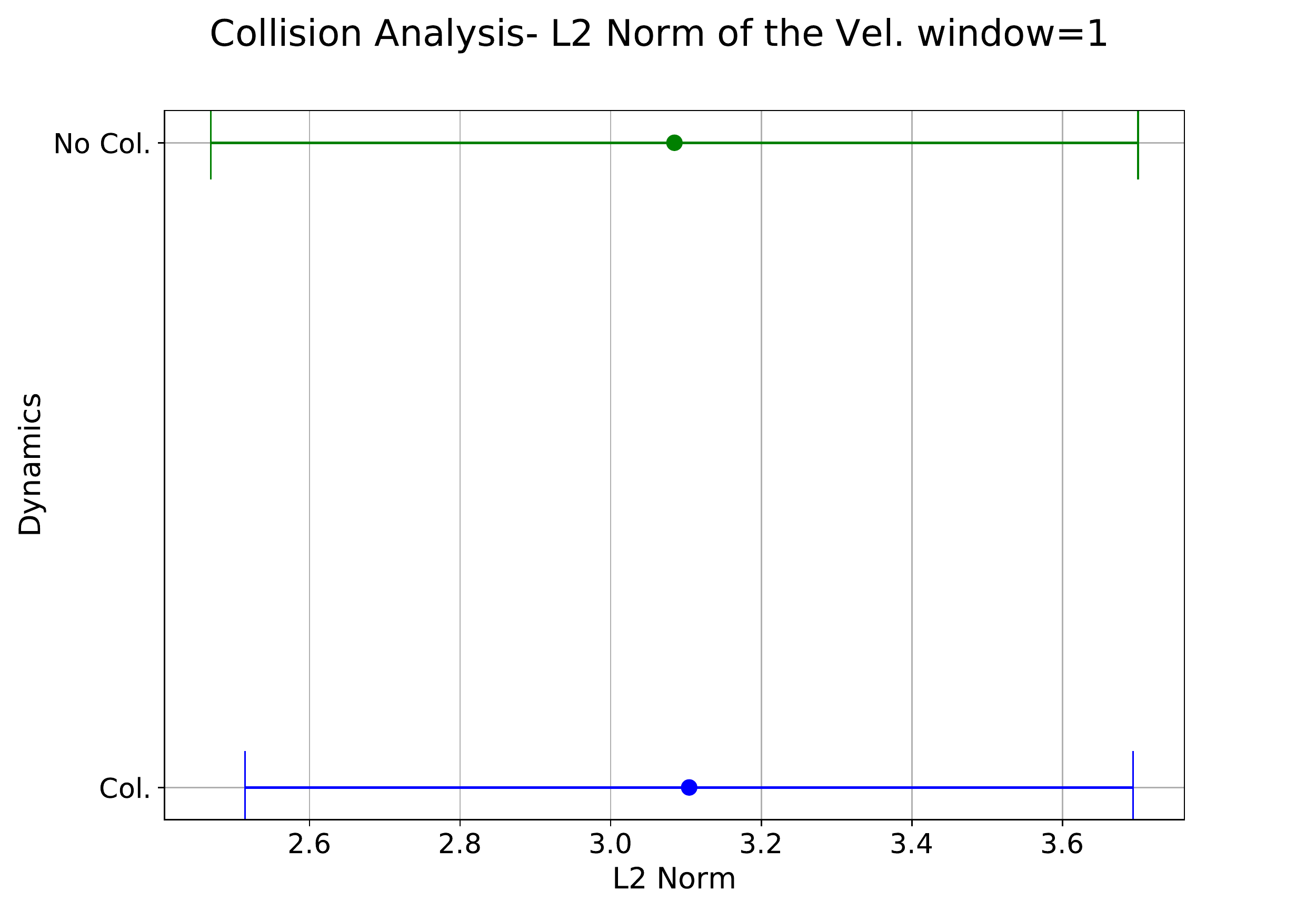}}%
\caption[Figure for the latent representation analysis of the baseline model for the projectile motion dataset.]{Mean and standard deviation values for the L2-norm of the latent acceleration and velocity for the projectile motion dataset. The figures display the values at the moment of collisions and other time steps.}
\label{fig:projectile_norm_spreads_all}
\end{figure}

We find out that the ODE2VAE baseline model could not capture the projectile motion dynamics with $a=2,3,5,7$ after it is trained for 300 epochs and converged. Therefore, we increased the number of latent units to 9. Although the model has increased predictive performance with $a=9$, it is not fully able to capture latent dynamics that resemble the real projectile motion.  In Figure \ref{fig:projectile_msepsnr}, we plot MSE and PSNR values for the test cases over the time steps. After time step four, there is an interval in which the ball hits the ground for each test case. Since the ball spends 0.1 second during its collision and the frames are separated for 0.1 second, the motion of the ball becomes stationary. The drop in the MSE values and the peak in the PSNR values may be because of the network's tendency to overfit when the ball becomes stationary. We present an example case in Figure \ref{fig:projectile_figure}. One of the challenges about the projectile motion is that there is a constant gravitational acceleration to learn. The results show that the model is not able to generate a constant acceleration field. On the other hand, there is a slight increment in the total kinetic energy when the ball is closer to the ground. In Figure \ref{fig:projectile_norm_spreads_all}, we present the norm of the acceleration and velocity latents at the time steps with and without collision. Although the true acceleration field is constant (except for the collision moments) due to the constant free-fall acceleration, the model is not able to learn a fixed acceleration field among the different test cases (see Figure \ref{fig:projectile_norm_spreads_acc}). We omit to comment on the norm of the latent velocity since the ground truth velocities are not the same in the test set. It is only reported for the sake of completeness.

\section{Discussion}
In this paper, we challenge the ODE2VAE model \cite{ode2vae} with synthetic physical motion datasets: bouncing balls, simple pendulum, and projectile motion. Since ODE2VAE infers latent dynamic trajectories by using coupled latent ODEs in a hierarchical latent space, it has a physics-guided inductive bias. We analyze the dynamical latent representations inferred by the model. Our experiments uncover the effects of the model's inductive bias over the inferred latent representations. It is shown that the ODE2VAE model is able to learn physically meaningful latent representations in an unsupervised setting. We observe that the model can learn physically plausible dynamic latent representations for the bouncing ball and simple pendulum datasets. Although the model generates successful predictions for the projectile motion dataset, its latent representations lack physical intuition. Additionally, our results empirically show that the uncertainty over the magnitude of the acceleration field increases during rare events and non-linear motions such as collisions.

Our work may be extended by challenging the ODE2VAE model with more complex motion datasets such as physical pendulum and double pendulum. Additionally, the latent acceleration field of the ODE2VAE model can be parameterized by using arbitrary Lagrangians or Hamiltonians \cite{cranmer2020lagrangian,hamiltonian_generative}, which may increase the effects of the inductive bias for learning physically plausible latent representations. Lastly, it is possible to check if ODE2VAE generates meaningful latent representations of the dynamical systems from other domains such as chemistry, biology, and medicine.
\section*{Acknowledgments}
This work is supported by Boğaziçi University Research Fund under the Grant Number 16903. We also thank Inzva for the computing resources provided.

\bibliographystyle{unsrt}  
\bibliography{templateArxiv}  

\newpage
\section*{Appendix}
\end{document}